\documentclass{article} 

\usepackage{amsmath}
\usepackage{fullpage}
\usepackage{url}
\usepackage{graphicx}
\usepackage{amssymb}
\usepackage{booktabs}       
\usepackage{xcolor}         
\usepackage{xspace}
\usepackage{float}
\usepackage{colortbl}
\usepackage{enumitem}
\usepackage{multirow}
\usepackage[mathscr]{eucal}
\usepackage[export]{adjustbox}
\usepackage{multicol,lipsum}
\usepackage{arydshln}
\usepackage{pifont}
\usepackage{bbding}
\usepackage[table, dvipsnames, svgnames]{xcolor}
\usepackage{comment}
\usepackage{wrapfig}
\usepackage{subcaption}
\usepackage{makecell}
\usepackage{pifont}
\usepackage[numbers]{natbib}

\def\W{{\bf W}}

\def\0{{\bf 0}}
\def\1{{\bf 1}}

\definecolor{darkgreen}{rgb}{0.3, 0.75, 0.3}
\definecolor{lightred}{rgb}{1.0, 0.5, 0.5}
\definecolor{col33}{RGB}{206, 239, 255}
\newcommand{\cmark}{\textcolor{darkgreen}{\checkmark}}%
\newcommand{\xmark}{\textcolor{lightred}{$\times$}}%

\definecolor{toysrusred}{RGB}{206, 17, 38}
\definecolor{toysRed}{RGB}{212, 48, 93}  
\definecolor{toysBlue}{RGB}{54, 90, 175}  
\definecolor{toysGreen}{RGB}{131, 192, 62}  
\definecolor{toysOrange}{RGB}{226, 139, 27}  
\definecolor{lightlightgrey}{rgb}{0.9,0.9,0.9}
\definecolor{cvprblue}{rgb}{0.21,0.49,0.74}

\newcommand*{\toysrus}{\textbf{\textcolor{toysRed}{T}\textcolor{toysOrange}{o}\textcolor{toysGreen}{y}\textcolor{toysRed}{s}\textcolor{toysBlue}{\reflectbox{R}}\textcolor{toysGreen}{U}\textcolor{toysRed}{s}}\@\xspace}

\newcommand*{\ourmethod}{Castoags\@\xspace}

\usepackage[pagebackref,breaklinks,colorlinks,allcolors=cvprblue]{hyperref}
\usepackage{stfloats}
\usepackage[capitalize]{cleveref}

\title{Revisiting Monocular In-Hand Object Scanning:\\ Dataset, Pose Estimation, and Surface Reconstruction}

\author{Di Wu  \quad Jeeeun Kim \quad Shu Kong \\
{\em website: \url{https://diwu96063.github.io/ToysRus/}}
\vspace{-3mm}
}
\date{}

\begin{document}
\maketitle

\begin{abstract}\noindent
In-Hand Object Scanning (IHOS) using commodity monocular smartphone cameras holds immense potential to democratize 3D digitization for non-expert users, accelerating applications in digital archiving, augmented reality, and custom fabrication.
Yet, IHOS remains largely understudied due to a lack of realistic datasets and systematic benchmarks.
To bridge this gap, we introduce \toysrus,
a high-quality dataset 
featuring diverse figurines captured in 4K UHD video 
using standard RGB smartphone cameras.
Analyzing monocular IHOS from first principles,
we examine structure-from-motion (SfM) for camera pose estimation, which is key to IHOS.
We identify two failure modes:
small baselines between consecutive frames, and sparse feature matches across long-range frames exacerbated by dynamic hand occlusions.
Guided by these insights, we propose a robust strategy that operates SfM on video sub-sequences sampled at sufficient angular intervals, and iteratively stitches the corresponding sub-trajectories into a unified global trajectory.
Finally, for high-fidelity surface reconstruction,
we introduce occlusion-aware 2D Gaussian Splatting which incorporates opacity rendering and explicitly mask hand regions in loss computation.
Quantitative and qualitative evaluations demonstrate that our framework significantly outperforms existing approaches,
including neural implicit representations and 3D foundation models,
in both geometric accuracy and photometric fidelity.

\end{abstract}

\begin{figure}[h]
\centering
\includegraphics[width=0.8\linewidth]{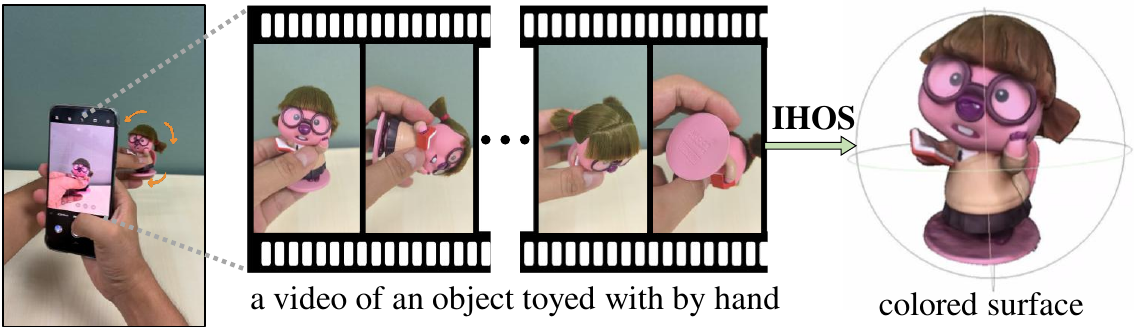}
\caption{\small
In-Hand Object Scanning (IHOS)
aims to reconstruct the complete, colored 3D surface of an object toyed with by hand.
Despite severe hand occlusions,
continuous interaction exposures all surface regions to the camera.
}
\label{fig:teaser}
\vspace{-1mm}
\end{figure}

\section{Introduction}

The widespread availability of smartphones offers an accessible avenue for democratizing 3D capture via In-Hand Object Scanning (IHOS), 
where a user ``\emph{toys}''\footnote{``\emph{Toying}'' refers to manipulating an object with fingers to expose all surface regions to a  camera, 
despite continuous dynamic hand occlusions. 
This contrasts a firm grasp, where the hands and object move rigidly together, permanently occluding certain parts of the object.}
with an object in front of a handheld camera (\cref{fig:teaser}).
Given an unconstrained video of a toyed object,
IHOS methods aim to reconstruct a complete, colored 3D surface model.
This capability enables diverse applications,
such as digitization of cultural heritage \cite{storeide2023standardization} and artworks \cite{bettio2013improving}, 
custom digital fabrication and 3D printing \cite{rindfleisch2019self, arbabian2020impact, berman2021howdiy, berman2020anyone, arabi2022mobiot, arabi2026eurexa},
and 
personal item retrieval in assistive robots and AR/VR wearables \cite{zhao2024instance, plizzari2024outlook, shen2025solving, plizzari2025spatial, shen2023high}.

IHOS has drawn growing attention \cite{huang2022reconstructing, wen2023bundlesdf, hampali2023hand, zhong2024color, jiang2025hand};
however, existing research remains constrained by synthetic or overly simplified capture setups and limited object variety.
For example,
most works rely on repurposed datasets like HOPE \cite{tyree20226} and HO3D \cite{hampali2020honnotate},
which are originally curated for robotic manipulation and 6DoF pose estimation.
These datasets contain low-resolution videos of common household items with simple geometries, such as milk cartons and pitchers.
Consequently,
they fail to satisfy the increasing demand for high-fidelity digitization and fabrication of complex, customized objects,
particularly in an era where modern smartphones routinely capture 4K UHD videos.
Using such datasets ultimately bottleneck methodology progress in IHOS.
To address these limitations,
we introduce a novel benchmark dataset called \toysrus, 
which features 4K UHD videos of diverse figurines with complex, intricate geometries.

The key to IHOS lies in accurately estimating per-frame camera poses,
which serve as a fundamental prerequisite for surface reconstruction methods such as
Marching Cubes \cite{lorensen1998marching}, 
Poisson surface reconstruction \cite{kazhdan2006poisson},
signed distance function (SDF) learning \cite{oechsle2021unisurf, mildenhall2021nerf, wang2021neus}, and Gaussian Splatting \cite{kerbl20233d, huang20242d}.
However, dynamic hand occlusions (\cref{fig:teaser}) severely complicate pose estimation \cite{hampali2023hand, swamy2023showme, jiang2025hand}, 
causing traditional pipelines such as Structure-from-Motion (SfM)~\cite{schonberger2016structure} to fail.
To address this issue,
some approaches \cite{huang2022reconstructing, jiang2024hand, swamy2023showme} constrain firm grasp and 
jointly estimate the hand and object pose.
However, maintaining static grip permanently occludes certain surface regions and leave them unable to be reliably reconstructed.
Alternatively, recent works \cite{jiang2025hand, chen2026forehoi} use generative AI agents to produce 3D object proxies and use them to assist in camera pose estimation and surface reconstruction.
While effective for common objects,
generating proxies fail on complex or customized items  (Fig.~\ref{fig:toysrus_result}),
significantly limiting their generalizability.

With the new dataset \toysrus,
we investigate IHOS from first principles by examining SfM for camera pose estimation.
We identify two failure modes: 
small baselines between consecutive frames  \cite{oliensis1999multi, vidal2002structure, bertel2019megaparallax, baker2020casualstereo}, and insufficient feature matches between long-range frames. 
We address these issues with a robust iterative strategy,
termed \textit{CamStich}, that performs SfM on video subsequences sampled at adequate angular intervals,
and then stitches the corresponding sub-trajectories into a coherent global trajectory.
We further introduce occlusion-aware 2D Gaussian Splatting (\textit{OA-2DGS}) to handle severe dynamic hand occlusions in colored surface reconstruction.
The developed methods lead to a simple pipeline for IHOS and constitute our technical insights and novelty.
We extensively evaluate our IHOS pipeline against existing approaches that exploit SDF learning for neural implicit surface reconstruction, and 3D foundation models.
Experiments demonstrate that ours significantly outperforms these approaches in terms of both geometric accuracy and photometric fidelity.

To summarize, we make three key contributions:
\begin{enumerate}[leftmargin=15pt, topsep=0pt, itemsep=5pt,parsep=-2pt]
    \item 
    We release a new dataset, \toysrus, featuring cellphone-captured 4K UHD videos of diverse figurines with complex geometries and fostering IHOS research.

    \item
    We rigorously analyze SfM from first principles on the 4K UHD videos,  
    revealing two issues caused by small baselines between consecutive frames and insufficient feature matches between long-range frames.

    \item
    We address the above issues with an iterative SfM and develop occlusion-aware 2D Gaussian Splatting for surface reconstruction.     
    Our two-step method significantly outperforms existing methods for IHOS.
\end{enumerate}

\section{Related Works}
\label{sec:related-works}

\textbf{Camera pose estimation for IHOS} is a crucial step and can intuitively be addressed by SfM.
However,
prior work points out that SfM fails under severe dynamic hand occlusions \cite{hampali2023hand, swamy2023showme, zhong2024color}.
To mitigate this issue,
some approaches leverage depth sensors 
\cite{wen2023bundlesdf, jiang2024hand} to enhance keypoint matching,
which limits their applicability to monocular setups.
Others \cite{huang2022reconstructing, cao2021reconstructing, hasson2019learning, ye2022s, fan2024hold, swamy2023showme} require a firm grasp and estimate hand pose to facilitate camera tracking; 
but static grasps permanently occlude certain surface regions, preventing reliable reconstruction.
Recent methods \cite{jiang2025hand, chen2026forehoi} employ generative AI agents to produce 3D proxies to aid pose estimation,
yet they fail on customized objects with complex geometries.
From first principles,
we re-examine SfM for camera pose estimation on IHOS videos and identify its failure modes.
We address these issues by introducing an iterative SfM strategy, demonstrating its effectiveness in achieving robust camera pose estimation even under severe dynamic hand occlusions.

\textbf{Colored surface reconstruction for IHOS},
given estimated camera poses,
can be achieved by a conventional pipeline that  first produces sparse 3D points triangulated from image features,
then generates dense point clouds using Multi-View Stereo (MVS)~\cite{schonberger2016pixelwise},
and lastly extracts surfaces via Martching Cubes \cite{lorensen1998marching} or screened Poisson surface reconstruction~\cite{kazhdan2013screened}.
However, 
sparse 3D points often lack sufficient structural information for reliable reconstruction.
To address this,
recent methods learn neural implicit functions to represent surfaces as the zero-level set of a signed distance function (SDF) \cite{zhong2024color, huang2022reconstructing, wen2023bundlesdf}.
These methods include 
IDR \cite{yariv2020multiview},
NeuS \cite{wang2021neus},
VolSDF \cite{yariv2021volume}, UNISURF \cite{oechsle2021unisurf}, and PET-NeuS \cite{wang2023pet}.
While effective,
optimizing an implicit function per object is time-consuming.
Alternatively, 
2D and 3D Gaussian Splatting (GS) \cite{kerbl20233d, guedon2024sugar, huang20242d}
represent 3D scenes with Gaussian primitives and utilize explicit ray-splat intersection to achieve high-quality, real-time surface reconstruction.
Yet, GS methods require accurate and dense view-dependent camera poses that are difficult to obtain from IHOS videos,
leaving them unexploited for this task.
Additionally, 
feed-forward modelsl like ForeHOI \cite{chen2026forehoi} train on large-scale synthetic datasets for efficient mesh prediction but generalize poorly to novel, complex geometries.
To overcome these limitations,
we present occlusion-aware 2DGS (OA-2DGS) by exploiting a simple loss term to handle dynamic occlusions in 2DGS,
achieving high computational efficiency and high-quality surface reconstruction.

{\bf Datasets for IHOS research}.
Existing IHOS research largely repurposes datasets like HOPE \cite{tyree20226} and HO3D \cite{hampali2020honnotate},
which were originally designed for robotic manipulation and object-hand pose estimation.
Consequently, 
methods developed on these datasets are biased toward tracking interaction dynamics \cite{cao2021reconstructing, hasson2019learning, ye2022s, fan2024hold, swamy2023showme} rather than recovering colored surfaces,
a task demanding high geometric accuracy and photometric fidelity.
As summarized in \Cref{tab:dataset},
legacy datasets exhibit severe limitations:
they feature low-resolution video instead of HD video, 
generic rather than customized objects,
constrained firm grasps instead of natural, user-friendly toying manipulation,
and static camera setups rather than convenient hand-held captures.  
Such characteristics fail to meet contemporary standards and the growing need for customized 3D capture and fabrication.
To bridge this gap and advance IHOS research,
we introduce the \toysrus dataset as a core contribution.

\textbf{Structure from Motion (SfM)} reconstructs 3D geometry and camera poses from multi-view images~\cite{schonberger2016structure, schonberger2016pixelwise}
through feature extraction \cite{lowe2004distinctive},
feature matching \cite{hartley2003multiple},
relative camera pose estimation \cite{seitz2006comparison},
outlier filtering \cite{fischler1981random},
and bundle adjustment \cite{triggs1999bundle}.
Toolboxes like COLMAP \cite{schonberger2016structure, schonberger2016pixelwise} integrate these components to support diverse  3D tasks,
while individual modules have been further advanced by learned feature extractors (e.g., SuperPoint~\cite{detone2018superpoint})
and matchers (e.g., SuperGlue~\cite{sarlin2020superglue} and LightGlue~\cite{lindenberger2023lightglue}).
Beyond the modular step-wise SfM pipeline,
recent 3D foundation models directly produce 3D reconstructions from 2D images, including VGGT~\cite{wang2025vggt}, MASt3R-SfM~\cite{duisterhof2025mast3r}, and MapAnything~\cite{keetha2025mapanything}.
Crucially, training these models relies on massive camera-pose annotations typically generated by COLMAP as pseudo ground truth.
Yet, 3D foundation models have not been explored for IHOS,
and we observe that they perform poorly in this setting.
Consequently, we re-examine SfM for camera pose estimation in IHOS videos featuring severe  dynamic hand occlusions,
identify its failure modes,
and propose targeted solutions that constitute our core technical novelty.

\section{\toysrus: Dataset and Evaluation Protocol}
\label{sec:dataset}

To foster research in IHOS,
we introduce the new dataset \toysrus, with the name reflecting the objective: ``you \emph{toy} with the object, I \emph{reconstruct} it''.

\subsection{Desiderata}
\toysrus reflects real-world IHOS scenarios
by addressing critical limitations in existing datasets (\Cref{tab:dataset}).
First, unlike HOPE \cite{tyree20226} and HO3D \cite{hampali2020honnotate}, which target robotic manipulation using lower-resolution RealSense RGB-D cameras (1080p/480p),
\toysrus provides 4K UHD videos at 60 fps captured by smartphone.
Second, 
prior datasets focus on common daily objects but
\toysrus features uncommon figurines to address the rising demand for customized object capture.
Third,
some datasets  \cite{huang2022reconstructing, hampali2020honnotate} 
enforce firm grasp of objects by hand, causing permanent occlusions and ignoring more natural interactions;
instead, \toysrus features user-friendly \emph{toying} manipulation (\cref{fig:teaser}).
Fourth,
unlike fixed-camera setups in existing datasets,
\toysrus employs hand-held smartphone capture under natural light, reflecting a practical, user-centric configuration.

\begin{figure*}[t]
\centering
\includegraphics[width=\linewidth]{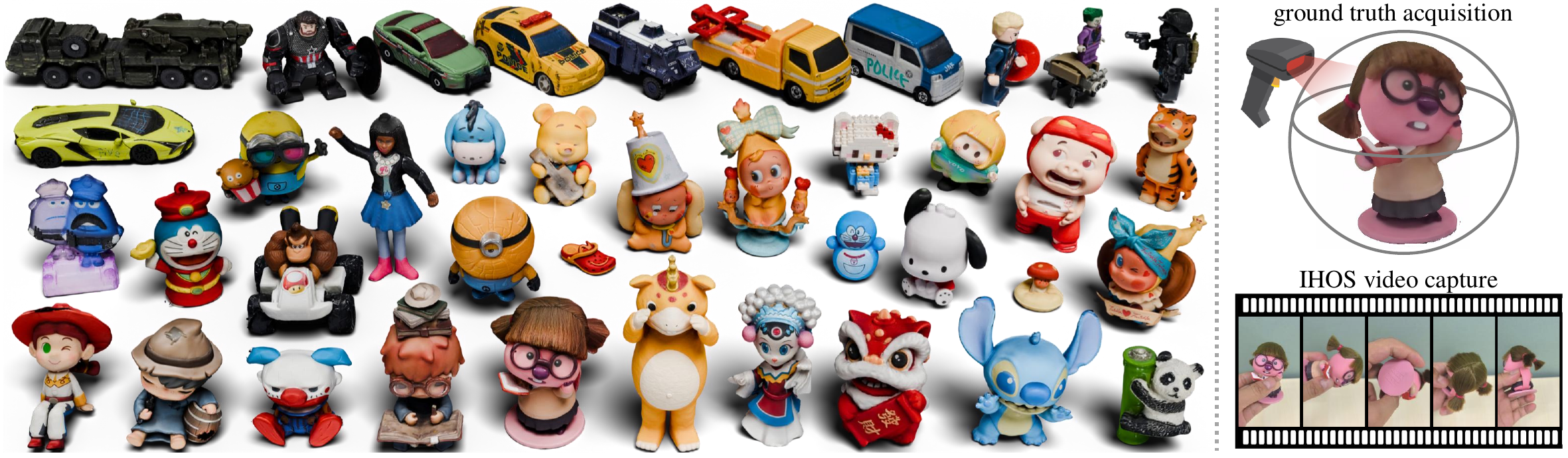}
\vspace{-7mm}
\caption{\small
{\bf The left panel} displays 
all the 40 figurines in the \toysrus dataset.
{\bf The right panel} illustrates  the acquisition of ground-truth colored surface by using a professional scanner, and a video of toying this figurine captured by a hand-held smartphone (\cref{fig:teaser}).
}
\label{fig:dataset}
\vspace{-1mm}
\end{figure*}

\begin{wraptable}{r}{0.55\linewidth}
\vspace{-5mm}
\small
    \centering
    \caption{\small
    \toysrus offers high-resolution, user-friendly 3D captures of complex geometries (\cref{fig:dataset}) via hand-held 4K UHD smartphone videos, addressing the limitations of existing datasets.
    }
    \label{tab:dataset}
    \vspace{-3mm}
    \setlength{\tabcolsep}{0.8mm}
    \resizebox{0.98\linewidth}{!}{
    \small
    \begin{tabular}{l c c c c c c c c c c c  c c c c c c}
    \toprule
        dataset 
        & task
        & firm-grasp 
        & customized
        & full video
        & \#objects  
        \\
        &   
        & vs. toying 
        & objects?
        & release?
        &
        \\
    \midrule

        HOD~\cite{huang2022reconstructing} 
        & {\color{darkgreen} IHOS} 
        & \color{lightred}firm
        & \color{darkgreen}Yes
        & \xmark
        & 35 
        \\ 
        
        HOPE~\cite{tyree20226} 
        & {\color{lightred}pose est.} 
        & \color{darkgreen}toying
        & \color{lightred}No
        & \xmark
        & 28 
        \\

        HO3D~\cite{hampali2020honnotate} 
        & {\color{lightred} pose est.} 
        & \color{lightred}\  firm
        & \color{lightred}No
        & \cmark
         & 10 
        \\
        
        IHO~\cite{zhong2024color} 
        & {\color{darkgreen}IHOS} 
        & \color{darkgreen}toying
        & \color{lightred}No
        & \cmark
        & 4 
        \\
        
        \toysrus 
        & {\color{darkgreen}IHOS} 
        & \color{darkgreen}toying
        & \color{darkgreen}Yes
        & \cmark
        & \textbf{40}
        \\
    \bottomrule
    \end{tabular}
    }
    \vspace{-3mm}
\end{wraptable}

\subsection{Dataset Collection}

\textbf{Ground truth acquisition.}
We acquire ground-truth mesh for each figurine by following the literature \cite{liu2022hoi4d} that uses multi-view photogrammetry software \cite{Realitycapture, Agisoft}.
We place a figurine on a turn-table under diffuse lighting and capture multi-view high-resolution images.
To ensure complete surface coverage, we reposition the figurine and repeat the above process.
We use the professional KIRI Engine \cite{kiriengine} to generate the 3D mesh for this figurine as its ground truth.

\textbf{Video collection.}
We use a commodity smartphone to capture 4K UHD videos (3,840$\times$2,160) for each figurine.
To reflect the user-friendly configuration,
we use one hand to hold the smartphone and the other to toy with a figurine (\cref{fig:teaser}) for $\sim$30 seconds.
In each video, we ensure that all surface regions of the figurine are exposed to the camera.

\subsection{Evaluation Protocol}
\label{ssec:evaluation_protocol}

{\bf Metrics.}
We assess methods using two types of metrics:
\begin{itemize}[leftmargin=15pt, topsep=0pt, itemsep=5pt,parsep=-2pt]
    \item 
        \emph{Geometric metrics} evaluate the geometric accuracy \cite{milef2024towards} by comparing the reconstructed surface with the ground truth.
        We adopt the Chamfer Distance, 
        Accuracy (the ratio of points sampled from the reconstructed surface that are true positives),
        Completeness (the ratio of points sampled from the ground-truth surface recalled by the reconstruction),
        and F1 score (the harmonic mean of Accuracy and Completeness).
    \item 
        \emph{Photometric metrics} focus on the reconstructed appearance.
        For a given camera pose,
        we render images from the ground-truth and the reconstructed colored surface, respectively,
        and compare them using
        the peak signal-to-noise ratio (PSNR), 
        the structural similarity index measure (SSIM) \cite{wang2004image} 
        and the LPIPS distance  \cite{zhang2018unreasonable}.
\end{itemize}

\textbf{Pre-processing.} 
To ensure fair comparisons,
we implement necessary pre-processing steps.
First, we normalize the ground-truth meshes to a unit scale before computing metrics,
ensuring metric comparability across objects of varying sizes.
Second, 
to evaluate geometric accuracy,
we align (unit-scale) reconstructed meshes with ground-truth using
Iterative Closest Point (ICP)~\cite{rusinkiewicz2001efficient}.
Third,
to evaluate photometric appearance,
we use albedo rendering to generate images for specific camera poses from the ground-truth and reconstructed meshes and compare their difference.

\begin{figure*}[t]
\centering
\vspace{-3mm}
\includegraphics[width=\linewidth]{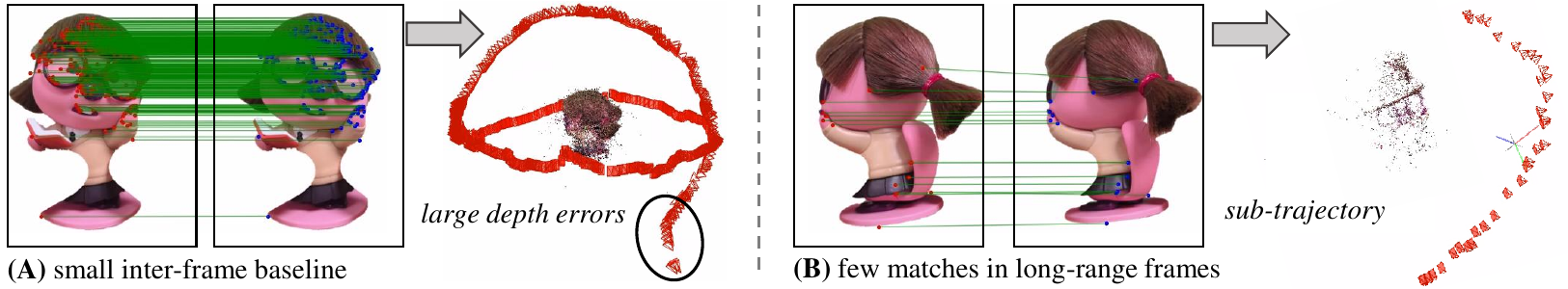}
\vspace{-5.0mm}
\caption{\small 
\textbf{A dilemma in pose estimation on an IHOS video.} 
{\bf (A)}: Leveraging dense frames suffers from the notorious small baseline issue \cite{oliensis1999multi, vidal2002structure, bertel2019megaparallax, baker2020casualstereo},
yielding erroneous depth estimates and camera trajectory drift.
{\bf (B)}: Sampling sparse frames ensures wide baseline but produces insufficient feature matches,
causing unreliable pose estimation, early termination, and incomplete reconstruction.
}
\label{fig:baseline_issue}
\vspace{-1mm}
\end{figure*}

\section{First-Principles Methodology Development}
\label{sec:method}

We approach IHOS from first principles by re-examining SfM for camera pose estimation:
why it struggles on IHOS videos (\S\ref{ssec:revisit-SfM}),
and how to make it work (\S\ref{ssec:subsequence_SfM} and \S\ref{ssec:CamPoseStitch}).
We further present occlusion-aware 2DGS to reconstruct colored surface (\S\ref{ssec:occlusion-aware-2DGS}).
\cref{fig:pipeline} illustrates our final method.

\subsection{Revisiting SfM Applied to IHOS Videos}
\label{ssec:revisit-SfM}

Existing works \cite{jiang2025hand, hampali2023hand} point out  the brittleness of SfM when applied to IHOS videos due to severe hand occlusions and dynamic object movements relative to the background scene.
A straightforward idea to address these issues is masking hand-associated pixels and background in each frame, focusing only on the target object.
Today's vision foundation models make these easily achievable.
We leverage the egocentric hand segmentation model \cite{zhang2022fine} and 
SAM2~\cite{ravi2024sam} to detect, segment and track hands and foreground object, and derive backgrounds in all video frames.
We execute SfM over the target object of the video frames with COLMAP.
Toward an in-depth analysis, 
we have extensively tuned hyperparameters with COLMAP and observe a dilemma (\cref{fig:baseline_issue}) related to the minimum triangulation angle when adding frames:
(1) a small minimum angle allows SfM to exploit large amounts of frames but produces erroneous depth estimates and eventually yields camera trajectory drift (\cref{fig:baseline_issue}A),
and (2) a large angle produces insufficient amounts of feature matches and thus makes SfM terminate early without exploiting enough frames (\cref{fig:baseline_issue}B).

\subsection{SfM on Subsequences of Video Frames}
\label{ssec:subsequence_SfM}

Motivated by the above dilemma, we develop an iterative variant of SfM, described as below.

\begin{figure*}[t]
\centering
\vspace{-3mm}
\includegraphics[width=\linewidth]{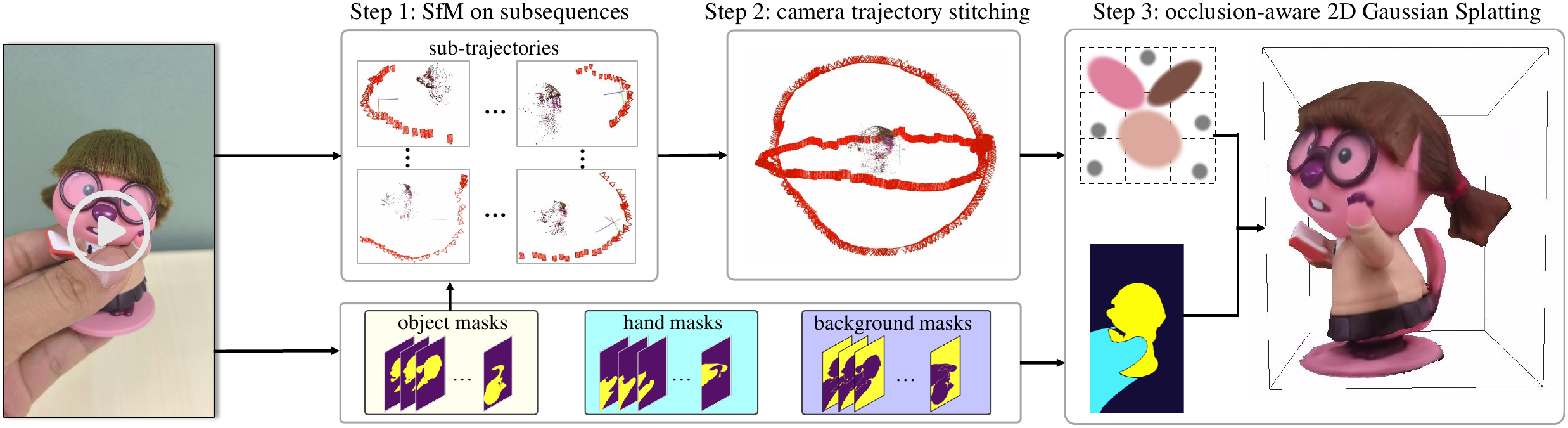}
\vspace{-6mm}
\caption{
    \textbf{Flowchart of the proposed method.} 
    Given an RGB video, 
    it segments and tracks the foreground object,
    background, 
    and hand.
    Then, 
    it performs SfM over subsequences of frames on the foreground object, resulting in multiple sub-trajectories of camera poses and point clouds (\S\ref{ssec:subsequence_SfM}).
    Next,
    it stitches them into a global camera trajectory and a dense point cloud (\S\ref{ssec:CamPoseStitch}, \cref{fig:stitch}).
    Lastly,
    it applies occlusion-aware 2DGS to reconstruct a colored surface (\S\ref{ssec:occlusion-aware-2DGS}).
    }
\label{fig:pipeline}
\vspace{-1mm}
\end{figure*}

\begin{figure}[b]
\centering
\vspace{-0mm}
\includegraphics[width=\linewidth]{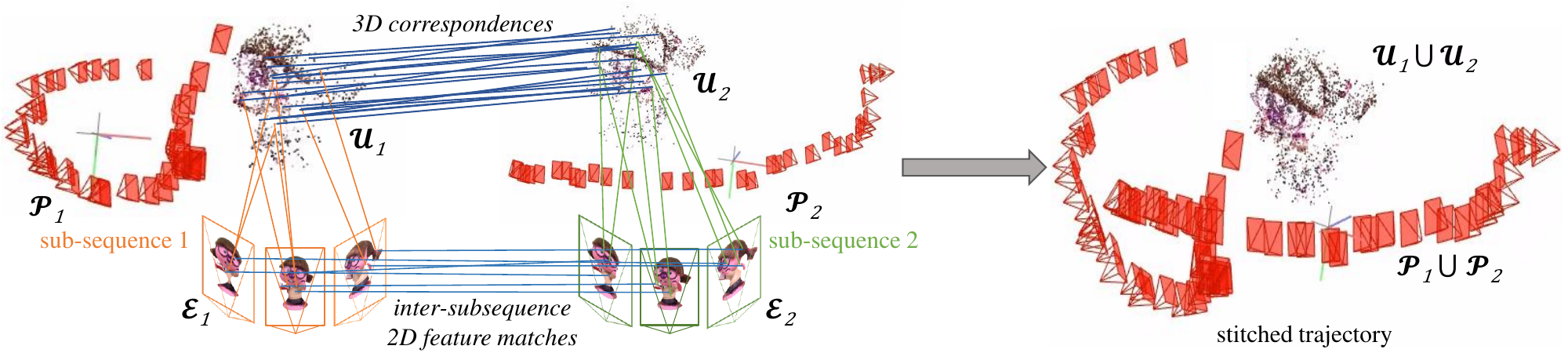}
\vspace{-7mm}
\caption{\small \textbf{Stitching sub-trajectories and point clouds.}
Although a pair of inter-subsequence frames may have too few 2D feature matches,
2D feature matches from more inter-subsequence frames and intra-subsequence 2D-to-3D correspondences (derived from SfM) collectively produce sufficient amounts of 3D correspondences.
With such,
we compute a Sim3 transformation matrix between the two point sets,
allowing to stitch sub-trajectories and point sets into a global trajectory and a dense cloud, respectively.
}
\label{fig:stitch}
\end{figure}

{\bf An iterative execution of SfM.}
Setting a proper minimum triangulation angle mitigates camera trajectory drift (\cref{fig:baseline_issue}A) but SfM terminates when insufficient feature matches are found between the exploited and remaining frames (\cref{fig:baseline_issue}B).
However, the remaining frames can still support another execution of SfM.
Hence, we propose to continually execute SfM over remaining frames.
Denote the set of remaining frames at iteration $k$ as $\mathcal{U}_k$;
$\mathcal{U}_1$ is the initial set of all frames.
Let $\mathcal{E}_{k}$ and $\mathcal{P}_k$ denote the sets of exploited frames and their corresponding camera poses estimated by SfM, respectively. 
The iterative SfM repeats the procedure:
$\mathcal{P}_{k}, \mathcal{E}_{k}, \mathcal{U}_{k+1} = \text{SfM}(\mathcal{U}_{k})$,
where $\mathcal{U}_k = \mathcal{E}_{k} \cup \mathcal{U}_{k+1}$.
The iteration terminates when no remaining frames can be exploited by SfM.
After each execution, SfM yields a sub-trajectory of camera poses and a set of 3D points (\cref{fig:stitch} and \cref{fig:pipeline}),
leaving a problem of stitching them into a global trajectory.
Before elaborating on our stitching technique (\S\ref{ssec:CamPoseStitch}), 
we present another variant of SfM below.

{\bf A parallel execution of SfM.}
From all frames of a given IHOS video, denoted by set $\mathcal{F}_{0}$,
we construct video subsequences, each of which is a subset of uniformly sampled frames.
Let $\mathcal{S}_{k}$ denote the $k^{th}$ subsequence, satisfying $\mathcal{F}_{0} = \cup_k \mathcal{S}_{k}$.
This SfM variant executes the procedure:
$\mathcal{P}_{k}, \mathcal{E}_{k}, \_ = \text{SfM}(\mathcal{S}_{k}) $.
Like the iterative SfM,
this variant also produces sub-trajectories,
for which a stitching method can form a global camera trajectory.

{\bf Remarks.}
The above two variants of SfM have respective advantages.
The iterative version requires a pre-defined minimum triangulation angle, allowing SfM to iteratively make use of all frames.
In contrast, 
the parallel version allows parallel processing and hence accelerated procession compared to the iterative one.
However,
it relies on a proper sampling interval as another hyperparameter,
setting which depends on the object motion speed.
For example, 
if a 30-second video at 60 fps (1,800 frames)
captures two full rotations ($720^\circ$),
sampling at every 5 frames amounts to an angular displacement of $\sim$2$^\circ$ between consecutive sampled frames.
This is a good triangulation angle for SfM in the COLMAP toolbox.
In fact,  we find that our final method \ourmethod is robust to different sampling intervals (refer to Suppl.~\S\ref{sec:tech_details_more}).
Hence, we apply the parallel variant of SfM  with sampling interval 5 in this work.

\subsection{Camera Sub-Trajectory Stitching}
\label{ssec:CamPoseStitch}

Over the sub-trajectories and the corresponding point clouds produced by SfM,
we stitch them into a global camera  trajectory and dense point cloud.
Over co-visible points across camera sub-trajectories,
we establish 3D correspondences by calculating an alignment transformation based on the insight (\cref{fig:stitch}):
\emph{while a pair of frames from two different sub-trajectories may contain too few feature matches,
more pairs of inter-subsequence frames and intra-subsequence 2D-to-3D correspondences collectively produce large amounts of 3D correspondences}.
With the 3D correspondences,
we adopt the well-established Umeyama \cite{umeyama1991least} algorithm to estimate a similarity transformation $\mathbf{T}$, which includes rotation $\mathbf{R}$, translation $\mathbf{t}$, and scale $s$. 
We apply $\mathbf{T}$ to align two sub-trajectories and point clouds (refer to Suppl.\S~\ref{sec:tech_details_more} for details).
Once stitching all sub-trajectories and point clouds, we perform a global Bundle Adjustment for the final refinement of camera poses and dense point cloud.
We name the pipeline of camera trajectory and point cloud generation as \textbf{CamStitch}.

\begin{figure*}[t]
    \centering
    {\tiny
        \makebox[0.10\textwidth][c]{\shortstack{ground-truth}}
        \makebox[0.08\textwidth][c]{\shortstack{Delaunay~\cite{labatut2009robust}}}
        \makebox[0.08\textwidth][c]{\shortstack{Poisson~\cite{kazhdan2013screened}}}
        \makebox[0.09\textwidth][c]{\shortstack{Color-NeuS~\cite{zhong2024color}}}
        \makebox[0.07\textwidth][c]{\shortstack{VGGT~\cite{wang2025vggt}}}
        \makebox[0.11\textwidth][c]{\shortstack{MASt3R-SfM~\cite{duisterhof2025mast3r}}}
        \makebox[0.11\textwidth][c]{\shortstack{MapAnything~\cite{keetha2025mapanything}}}
        \makebox[0.08\textwidth][c]{\shortstack{MAtCha~\cite{guedon2025matcha}}}
        \makebox[0.08\textwidth][c]{\shortstack{Dynhor~\cite{jiang2025hand}}}
        \makebox[0.08\textwidth][c]{\shortstack{ForeHOI~\cite{chen2026forehoi}}}
        \makebox[0.06\textwidth][c]{\shortstack{\bf\ourmethod}}
    }
    
    \includegraphics[width=\textwidth]{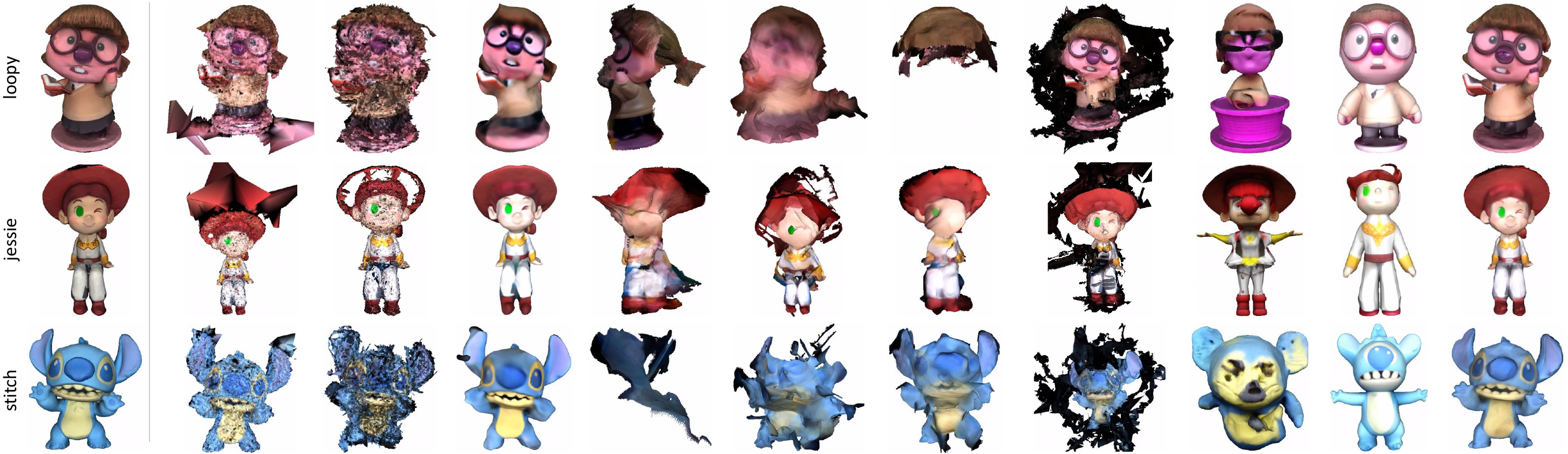} 
    \vspace{-6mm}
    \caption{\textbf{Visual results on \toysrus.} 
    Among all the compared methods,
    \ourmethod produces the most visually high-fidelity surface reconstructions.
    Refer to \Cref{tab:toysrus_result} for quantitative comparisons,
    and Suppl. \cref{suppl_fig:toysrus_result_more} for more qualitative comparisons.
    }
    \label{fig:toysrus_result}
    \vspace{-1mm}
\end{figure*}

\begin{table}[t]
\setlength{\tabcolsep}{0.5mm}
    \centering
    \caption{
    \small
    \textbf{Benchmarking on \toysrus.} 
    Our \ourmethod runs CamStitch for SfM and OA-2DGS for surface reconstruction.
    It performs the best w.r.t both geometric accuracy and photometric fidelity under moderate inference time cost (in hours).
    Yet, modern 3D foundation models yields poor performance for IHOS.
    }
    
    \vspace{-3mm}
    \resizebox{0.75\linewidth}{!}{
    \begin{tabular}{l>{\small}crrrrrrrr}
        \toprule
         & & \multicolumn{4}{c}{Geometry} & \multicolumn{3}{c}{Rendering} \\
        \cmidrule(lr){3-6}
        \cmidrule(lr){7-9}
         method
         & time $\downarrow$ 
         & acc.$\uparrow$ 
         & compl.$\uparrow$ 
         & F1$\uparrow$ 
         & CD$\downarrow$ 
         & PSNR$\uparrow$ 
         & SSIM$\uparrow$
         & LPIPS$\downarrow$ 
         \\
        \midrule

        \rowcolor{lightlightgrey}\multicolumn{10}{l}{SfM via default COLMAP \& dedicated methods for surface construction.} \\
        Delaunay \cite{labatut2009robust} $_\text{\color{gray}CGF'09}$ & 10.0 & 67.98 & 83.83 & 74.53 & 86.11 & 15.53 & 0.56 & 0.48 & \\
        
        Poisson \cite{kazhdan2013screened} $_\text{\color{gray}TOG'13}$ & 10.0 & 67.40 & 92.64 & 76.76 & 3.29 & 15.25 & 0.52 & 0.53 \\

        Color-NeuS \cite{zhong2024color} $_\text{\color{gray}3DV'24}$ & 12.0 & 82.88 & 80.67 & 81.75 & 2.24 & 14.16 & 0.65 & 0.39 \\

        \midrule
        \rowcolor{lightlightgrey}\multicolumn{10}{l}{SfM via foundational models \& our OA-2DGS for surface reconstruction} \\
        VGGT \cite{wang2025vggt} $_\text{\color{gray}CVPR'25}$ & 0.2 & 23.90 & 19.17 & 19.74 & 138.22 & 7.27 & 0.21 & 0.89 \\
        
        MASt3R-SfM \cite{duisterhof2025mast3r} $_\text{\color{gray}3DV'25}$ & 1.0 & 20.42 & 29.85 & 23.06 & 114.85 & 12.44 & 0.58 & 0.54 \\

        MapAnything \cite{keetha2025mapanything} $_\text{\color{gray}3DV'26}$ & 0.2 & 36.80 & 40.54 & 35.52 & 97.07 & 10.34 & 0.47 & 0.67 \\

        \midrule
        \rowcolor{lightlightgrey}\multicolumn{10}{l}{Direct surface reconstruction via modern 3D foundation models} \\
        MAtCha \cite{guedon2025matcha} $_\text{\color{gray}CVPR'25}$ & 0.3 & 22.11 & 56.10 & 31.49 & 137.35 & 15.43 & 0.67 & 0.44 \\
        
        Dynhor \cite{jiang2025hand} $_\text{\color{gray}CVPR'25}$ & 9.0 & 28.83 & 28.84 & 28.46 & 45.81 & 6.33 & 0.15 & 0.97 \\

        ForeHOI \cite{chen2026forehoi} $_\text{\color{gray}CVPR'26}$ & 0.1 & 29.58 & 28.34 & 28.51 & 57.14 & 5.67 & 0.09 & 1.00 \\

        \midrule
        
        \rowcolor{col33}{\bf \ourmethod} $_\text{  {\color{gray}ours}}$  & 0.5 &  \textbf{94.23} & \textbf{93.39} & \textbf{93.80} & \textbf{0.74} & \textbf{17.68} & \textbf{0.75} & \textbf{0.29} \\
        \bottomrule
    \end{tabular}}
    \label{tab:toysrus_result}
    \vspace{-1mm}
\end{table}

\subsection{Occlusion-Aware Surface Reconstruction}
\label{ssec:occlusion-aware-2DGS}
The recovered global trajectory and point cloud enables surface reconstruction with various methods such as the classic methods Delaunay \cite{labatut2009robust}  and Poisson \cite{kazhdan2013screened}.
Beyond these traditional methods, we propose occlusion-aware 2D Gaussian Splatting (OA-2DGS) for surface reconstruction.
Note that 2DGS \cite{huang20242d} excels in surface reconstruction for static scenes but does not handle dynamic occlusions (caused by hand in IHOS).
It uses a reconstruction loss term between a frame ${\mathbf{I}}$ and its rendered frame $\hat{\mathbf{I}}$.
We enhance it by adding a loss term to handle dynamic hand occlusions.
Recall that, for each frame ${\mathbf{I}}$, 
we derive binary masks for hand, foreground object, and background (\cref{fig:pipeline}).
For this frame, we define its target opacity ${\mathbf{A}}$ as the foreground mask, i.e., setting the foreground to 1 and elsewhere to 0.
We design an $L_1$ term  between ${\mathbf{A}}$ and the 2DGS rendered opacity $\hat{\mathbf{A}}$.
Within this term, we further apply a per-pixel weight $\W$ to ignore any reconstructions on the hand regions. 
The weight $\W$ is the complement of the hand mask, with hand regions set to 0 and elsewhere to 1.
Collectively, our designed term is $\Vert \mathbf{W} \odot (\hat{\mathbf{A}} - \mathbf{A}) \Vert_1$,
which focuses on reconstructing the foreground object without being distracted by the background and hand regions and does not compute opacity predictions on hand regions. 
Below is our final loss OA-2DGS:
\begin{equation}
    \mathcal{L}_c = (1 - \lambda) \big( \Vert \mathbf{W} \odot (\hat{\mathbf{I}} - \mathbf{I}) \Vert_1 + \Vert \mathbf{W} \odot (\hat{\mathbf{A}} - \mathbf{A}) \Vert_1 \big) 
    + \lambda \mathcal{L}_{\text{D-SSIM}}(\mathbf{W} \odot \hat{\mathbf{I}}, \mathbf{W} \odot \mathbf{I}),
\label{eq:final_loss}
\end{equation}
where $\lambda=0.2$ by default \cite{huang20242d}.
$\mathcal{L}_c$ removes the effect of dynamic hand,
helps learning opacity pertaining to the foreground object,
and enables the follow-up Marching Cubes to produce clean and smooth surface.
To train 2DGS, we use $\mathcal{L}_c$ along with standard regularizers, including the distance between the ray-splat intersections and splats' normal with the gradients of the depth maps.
We name this method as 
Occlusion-Aware 2D Gaussian Splatting ({\bf OA-2DGS}).

\section{Experiments}

With extensive experiments, we validate our method \ourmethod.
We start with the compared methods, datasets, and important implementation details,
followed by benchmarking results and ablation studies.

\begin{wrapfigure}{r}{0.50\linewidth}
\centering
\vspace{-4mm}
\begin{minipage}[c]{0.05\linewidth}
    \centering
    \rotatebox{90}{\scriptsize \makebox[1cm][c]{reference}} \\ \vspace{0.2cm}
    \rotatebox{90}{\scriptsize \makebox[1.4cm][c]{ground-truth}} \\ \vspace{0.3cm}
    \rotatebox{90}{\scriptsize \makebox[1.5cm][c]{Color-NeuS \cite{zhong2024color}}} \\ \vspace{0.2cm}
    \rotatebox{90}{\scriptsize \makebox[1.3cm][c]{\bf \ourmethod}}
\end{minipage}\hfill
\begin{minipage}[c]{0.95\linewidth}
    \includegraphics[width=\linewidth, clip=true,trim = 0mm 0mm 0mm 0mm]{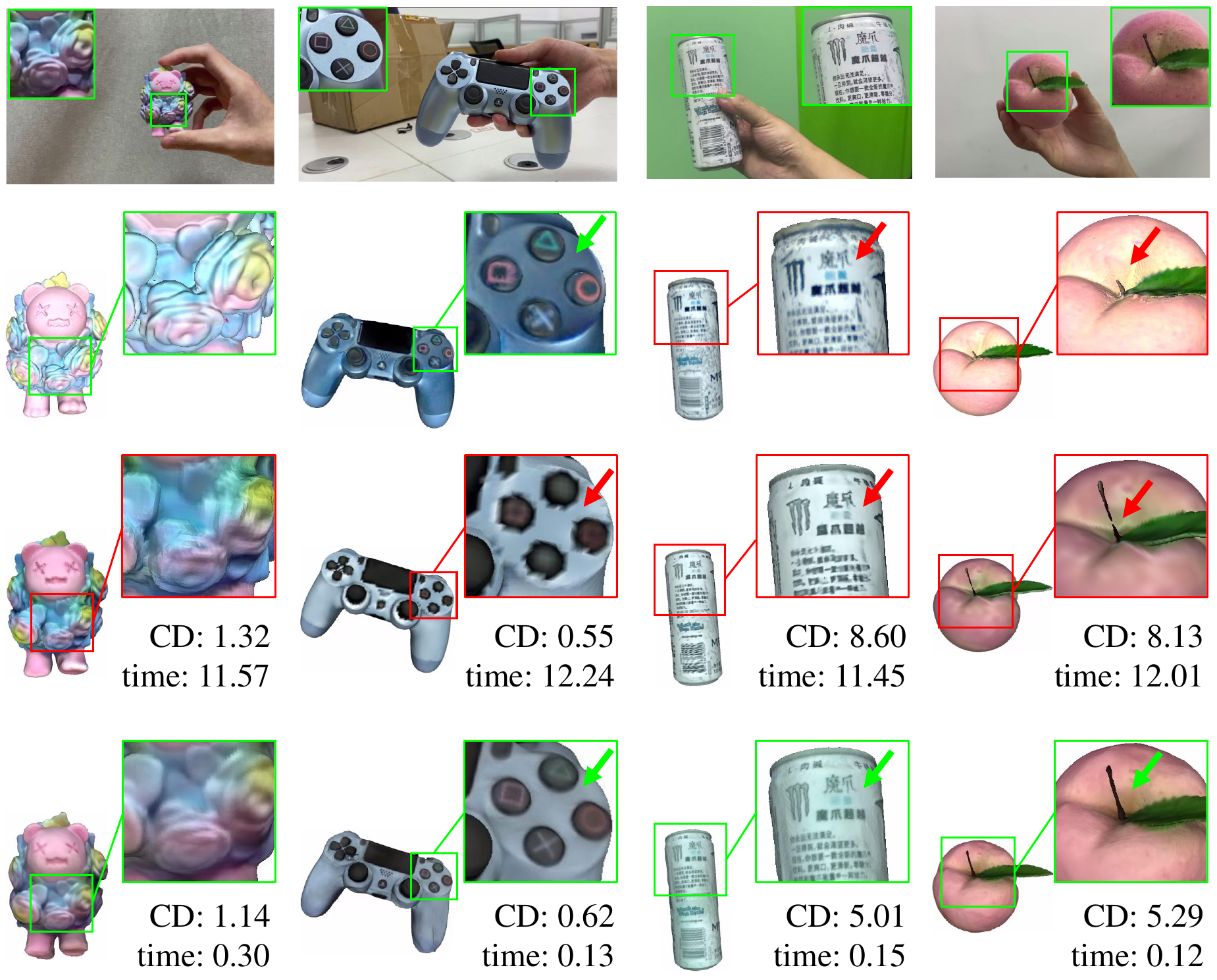}
\end{minipage}
\vspace{-3mm}
\caption{\small
    \textbf{Benchmarking on IHO} w.r.t Chamfer Distance (CD) and time cost.
    Red boxes highlight unsatisfactory reconstructions by Color-NeuS or imperfect ground truth.
    \ourmethod outperforms Color-NeuS, which is the most competitive method (Table~\ref{tab:toysrus_result}).
    However, Color-NeuS takes 12 hours to reconstruct one figurine, whereas \ourmethod takes 20 minutes.
    Refer to Suppl.~\S\ref{sec:more_results_more_datasets} for more metrics. 
    Zoom-in for better visualization.
}
\vspace{-4mm}
\label{fig:iho_result}
\end{wrapfigure}

\textbf{Methods.}
We compare three categories of methods:
\begin{itemize}[leftmargin=10pt, topsep=0pt, itemsep=5pt,parsep=-2pt]
    \item
        \emph{SfM via COLMAP + methods for surface construction}.
        Delaunay \cite{labatut2009robust} and Poisson \cite{kazhdan2013screened}  reconstruct surfaces from point clouds produced by SfM with COLMAP with default settings.
        Color-NeuS~\cite{zhong2024color} also relies on COLMAP for camera pose estimation but it trains an implicit SDF to reconstruct surface for each object.

    \item
        \emph{SfM via foundational models + our OA-2DGS for surface reconstruction}.
        MASt3R-SfM~\cite{duisterhof2025mast3r} outputs 3D points and uses them to estimate camera poses;
        VGGT~\cite{wang2025vggt} and       
        MapAnything~\cite{keetha2025mapanything} generate 3D points and estimate camera poses.
        As they do not reconstruct surfaces,
        we run our OA-2DGS over their output point clouds and camera pose estimates to reconstruct surfaces.

    \item
        \emph{Direct surface reconstruction via modern 3D foundation models}.
        MAtCha~\cite{guedon2025matcha} directly outputs mesh by taking as input frames.
        Dynhor \cite{jiang2025hand} uses AI agents to produce text descriptions for a target object, exploits a text-to-3D model to produce a 3D proxy.
        It uses this proxy to initialize camera pose for joint optimization and SDF learning.
        As Dynhor does not release code and its adopted models were deprecated, 
        we re-implement it using GPT-5.2 \cite{singh2025openai} and the recent text-to-3D model TRELLIS \cite{Xiang_2025_CVPR}.
        ForeHOI \cite{chen2026forehoi} directly trains a feed-forward model to reconstruct surfaces from input frames.
\end{itemize}
Notably,
as most of the methods do not handle dynamic hand occlusions,
we mask hand and background pixels, 
like how we do for our \ourmethod in video frames,
as their input.

\textbf{Datasets and Metrics.}
Other than \toysrus, we use IHO \cite{zhong2024color} in experiments.
IHO contains four videos, each for a single hand-held object.
It provides ground-truth camera pose, allowing evaluating camera pose estimation of our CamStitch,
e.g., using metrics mean reprojection error (mRE), absolute translation error (ATE), and relative rotation error (RPE$_r$), and relative translation error (RPE$_t$).
Suppl. \S\ref{sec:more_results_more_datasets} provides results on more datasets.

{\bf Implementation details.}
All the methods run on a computer with an Intel Core i7-13700 CPU and an 
NVIDIA  4090 GPU.
When memory allows, 
the compared methods exploit entire videos.
Yet,
3D foundation models eventually suffer from the memory bottleneck, except MASt3R-SfM which process frames sequentially.
In this case, we 
we uniformly sample video frames to maximize GPU utilization for VGGT (32),
MapAnything  (32), 
and MAtCha (16).

\subsection{Benchmarking Results}
\label{ssec:benchmark}

\Cref{tab:toysrus_result} and \Cref{fig:toysrus_result,fig:iho_result}
list numeric metrics and visual comparisons of all the methods.
Our \ourmethod significantly outperforms the compared methods.
Somewhat surprisingly,
methods based on 3D foundation models (e.g., VGGT and MAtCha) achieve poor performance,
an observation aligning with 
\cite{chen2026forehoi} which reports that such models fail to provide a good SfM initialization in face of dynamic occlusions.
Dynhor and ForeHOI, specifically designed for IHOS, struggle on \toysrus.
Their failures are rooted in a common limitation:
both propose producing 3D priors using generative AI agents which do  not generalize to custom, novel objects as featured in \toysrus.

Notably,
our \ourmethod takes only 0.5 hours for surface reconstruction per figurine (Table~\ref{tab:toysrus_result}),
whereas 
the runner-up method Color-NeuS takes 12 hours and 
Dynhor~\cite{jiang2025hand} takes 9 hours. 
Moreover,
methods based on 3D foundation models are much faster owing to their feed-forward mechanisms:
for camera pose estimation,
MapAnything and VGGT takes seconds and  MASt3R-SfM takes 48 minutes;
they further take 0.2 hours to reconstruct surfaces using OA-2DGS.
MAtCha takes 0.3 hours to directly output reconstructed surfaces,
and ForeHOI is even faster that outputs surfaces in 0.1 hours. 
However,
these methods yield poor quality of reconstructed surfaces,
since they have not learned similar complex geometries as the custom figurines and hence have limited generalizability.

\begin{figure*}[t]
    \centering
    \begin{minipage}[t]{0.48\textwidth}
        \vspace{0pt}
        \centering
        \includegraphics[width=\linewidth]{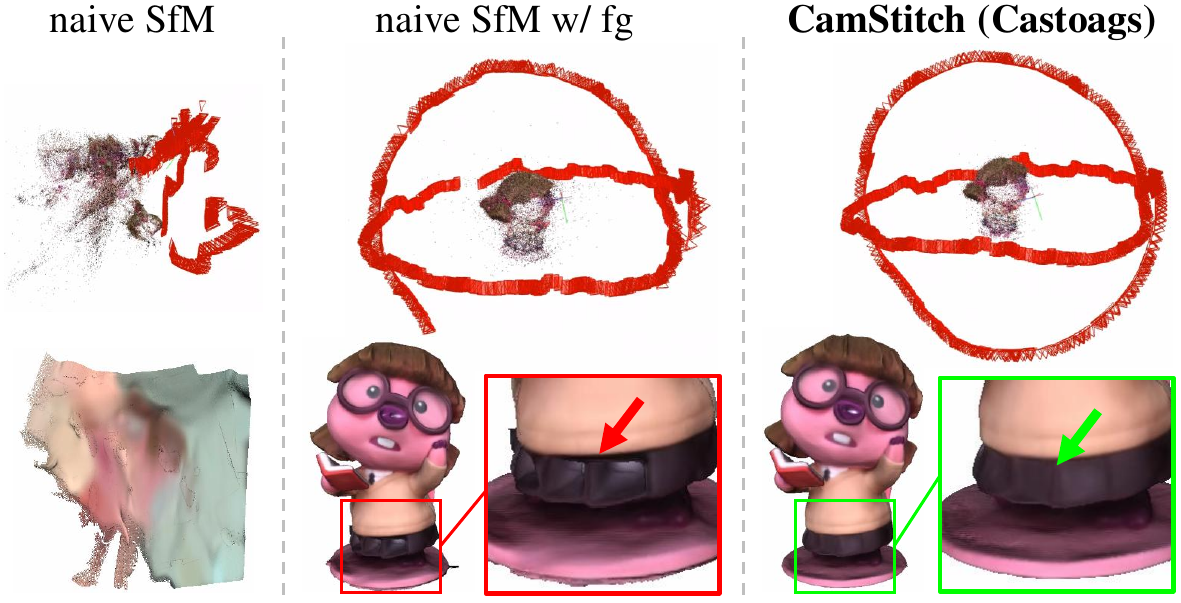}
    \end{minipage}
    \hfill
    \begin{minipage}[t]{0.5\textwidth}
        \vspace{0pt}
        \makeatletter\def\@captype{table}\makeatother
        \centering
        \caption{\small \textbf{Ablation of CamStitch} on \toysrus video.
        We implement the parallel variant of SfM with COLMAP, followed by our OA-2DGS for surface reconstruction.
        Our method \ourmethod
        drastically reduces processing time and significantly improves surface reconstruction.
        }
        \label{tab:COLMAP-analysis}
        \vspace{-2mm}
        \setlength{\tabcolsep}{1mm}
        \resizebox{\linewidth}{!}{
        \begin{tabular}{l rr rrrrr}
            \toprule
             \multirow{2}{*}{method} & \multicolumn{2}{c}{SfM} & \multicolumn{5}{c}{Surface Reconstruction}  \\
             \cmidrule(lr){2-3} \cmidrule(lr){4-8}
             & mRE$\downarrow$ & time$\downarrow$ & acc.$\uparrow$ & compl.$\uparrow$ & F1 $\uparrow$ & CD $\downarrow$ & time $\downarrow$\\
            \midrule
            naive SfM & 1.30 & 6.0 & 0.82 & 6.27 & 1.46 & 6.86k & 0.17 \\
            \quad w/ fg & 0.77 & 2.0 & 94.65 & 96.31 & 95.47 & 0.50 & 0.17 \\
            \rowcolor{col33}
            \quad\quad \textbf{CamStitch} & \textbf{0.51} & \textbf{0.3} & \textbf{97.84} & \textbf{97.96} & \textbf{97.90} & \textbf{0.36} & \textbf{0.15}\\
            \bottomrule
        \end{tabular}}
    \end{minipage}

    \vspace{-2mm}

    \caption{\small \textbf{Visual results} with progressive modifications on SfM (\Cref{tab:COLMAP-analysis}).
    Naive SfM, operating on raw video frames, fails to recover camera poses. 
    When focusing only on target objects, 
    SfM improves but produces incomplete camera trajectories.
    CamStitch (a crucial component in \ourmethod) recovers the full camera trajectory and helps reconstruct high-quality mesh.
    }
    \label{fig:COLMAP-analysis}
\end{figure*}

\subsection{Analysis: Structure from Motion by CamStitch}
\label{subsec:COLMAP-analysis}

Recall that SfM \cite{schonberger2016structure} struggles to accurately estimate camera poses on video frames with severe dynamic hand occlusions (\cref{fig:baseline_issue}).
As shown in \Cref{tab:COLMAP-analysis},
it stalls for 6 hours to terminate and yields poor estimates of camera poses.
We now run it over foreground pixels (i.e., masking hand and background pixels), dubbed ``w/ fg''.
This helps SfM exploit more frames and estimate camera poses more accurately.
Yet, it still requires 2 hours to complete camera pose estimation.
Importantly, 
adopting proposed parallel SfM followed by stitching sub-trajectories of estimated camera poses, i.e., CamStitch,
significantly accelerates pose estimation and produce remarkably better pose estimates, eventually leading to the best surface reconstruction. 
\cref{fig:COLMAP-analysis} compares visual results of the three methods.

\begin{wraptable}{r}{0.48\linewidth}
    \centering
    \vspace{-5mm}
    \caption{\small
    \textbf{CamStitch improves pose estimation}
    over SfM.
    We report averaged results over the objects of the IHO dataset w.r.t  mRE (px), ATE (m), RPE$_t$ (m), RPE$_r$ ($^\circ$), and time (hrs).
    }
    \label{tab:campose_estimation}
    \vspace{-2mm}
    \setlength{\tabcolsep}{1.8mm}
    \resizebox{\linewidth}{!}{
    \begin{tabular}{lrrrrr}
        \toprule
        method 
        & mRE$\downarrow$ & ATE$\downarrow$ & RPE$_r$$\downarrow$ & RPE$_t$$\downarrow$ & time$\downarrow$ \\
        \midrule
        SfM w/ fg & 1.11 & 0.48 & 0.45 & 1.80 & 3.25 \\
        \rowcolor{col33}{\textbf{CamStitch}} & \textbf{0.54} & \textbf{0.10} & \textbf{0.10} & \textbf{1.73} & \textbf{0.44} \\
        \bottomrule
    \end{tabular}}
    \vspace{-5mm}
\end{wraptable}
We compare the precision of estimated camera 
poses on the IHO dataset, which provide ground-truth
camera poses for video frames.
As shown in \Cref{tab:campose_estimation},
our CamStitch achieves more precise camera pose estimation than SfM w/ fg, with one order less computational time cost.

\begin{figure*}[t]
\vspace{-3mm}
    \includegraphics[width=0.98\linewidth]{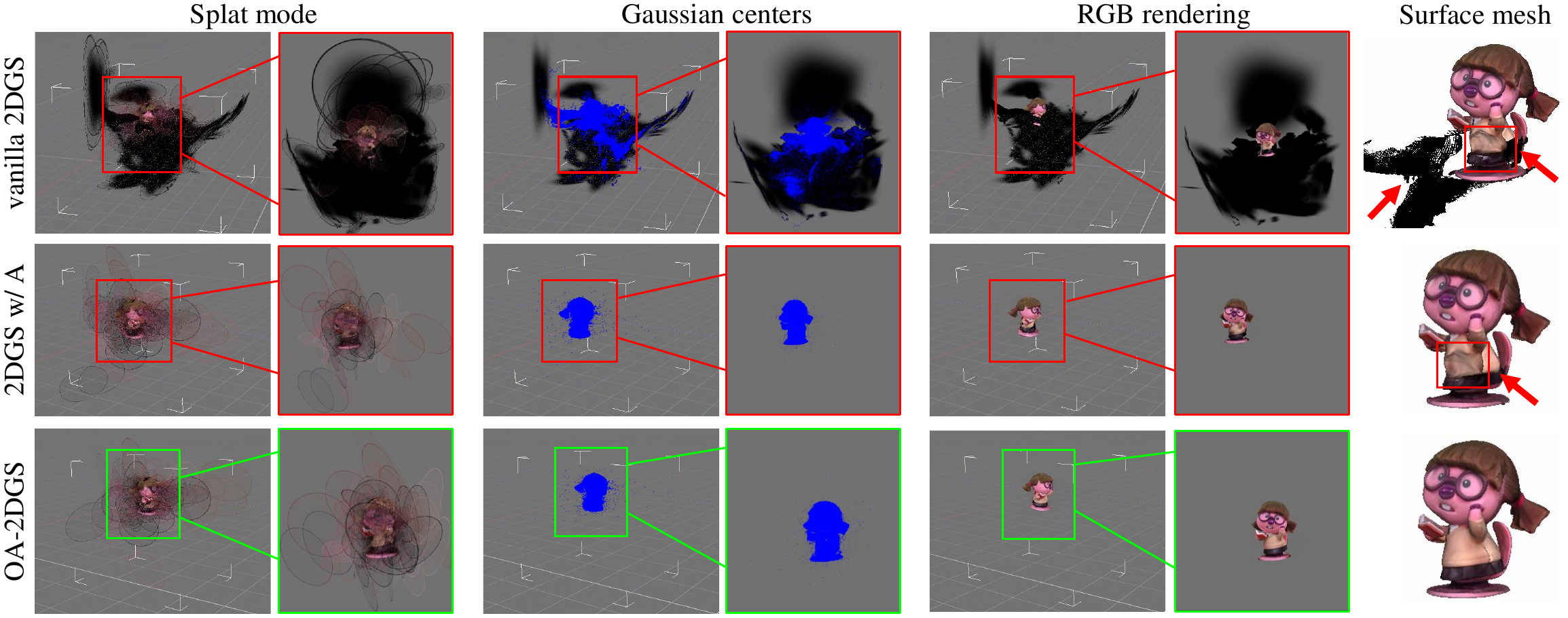} 
    \vspace{-3mm}
    \caption{
        \small
        \textbf{Qualitative comparisons.}
        We visualize the reconstruction results using splat mode, Gaussian centers, RGB rendering, and the final reconstructed surface mesh.
       Our full OA-2DGS pipeline (w/ A, W) effectively integrates opacity supervision with the hand occlusion mask, proving essential for extracting clean, high-fidelity object geometry under severe hand occlusions.
    }
    \label{fig:oa_2dgs_ablation}
    \vspace{-1mm}
\end{figure*}

\subsection{Analysis: Surface Reconstruction by OA-2DGS}
\label{subsec:OA-2DGS-analysis}

\begin{wraptable}{r}{0.45\linewidth}
\setlength{\tabcolsep}{2.5mm}
\vspace{-5mm}
    \centering
    \caption{\textbf{Ablation study for OA-2DGS} w.r.t 
    loss terms.
    The vanilla 2DGS, in the top row,
    does not apply opacity loss term nor the per-pixel weight.
    It produces poor surface reconstruction results; refer to the first row of \cref{fig:oa_2dgs_ablation}.
    Adding an opacity loss term significantly enhances the quality of reconstructed surface, while our final method OA-2DGS (cf. last row)
    yields the best metrics.
    Refer to \cref{fig:oa_2dgs_ablation} for visual comparisons.
    }
    \vspace{-3mm}
    \resizebox{0.98\linewidth}{!}{
    \begin{tabular}{cc cccr}
        \toprule
         $\mathbf{A}$ & $\mathbf{W}$ & acc.$\uparrow$ & compl.$\uparrow$ & F1 $\uparrow$ & CD $\downarrow$ \\
        \midrule

        N/A & N/A 
        & 48.95 
        & 67.39 
        & 56.71 
        & 114.59 \\

        Yes & N/A
         & 88.42
         & 90.80
         & 63.99
         & 1.86 \\
        
        \rowcolor{col33}
        Yes & Yes
         & \textbf{96.16}
         & \textbf{96.47}
         & \textbf{96.31}
         & \textbf{0.63} \\
        \bottomrule
    \end{tabular}}
    \label{tab:oa_2dgs_ablation}
    \vspace{-3mm}
\end{wraptable}
We analyze the proposed OA-2DGS for surface reconstruction.
Here, we use the global camera pose trajectory and point cloud obtained by the precedent step CamStitch.
First, we apply the vanilla 2DGS directly on the foreground pixels of video frames, removing hand and background regions.
As shown in \cref{fig:oa_2dgs_ablation},
the  vanilla 2DGS produces surface meshes with severe floaters.
Next,
we apply the opacity loss term without the per-pixel weight, termed ``2DGS w/ A''.
This approach eliminates a great amount of floaters,
yielding a cleaner surface mesh, as shown in the middle row of \cref{fig:oa_2dgs_ablation}. 
However,
the reconstruct surfaces have notable artifact around hand-occluded regions,
due to the mechanism of this approach that treats the hand occluded regions as transparent regions and compute opacity losses on them. 
As a result, 2DGS learns to erroneously eliminate these hand-occluded regions along with true background, introducing artifacts on the object surface under the occlusion.
Finally,
we additionally apply the per-pixel weight $\mathbf{W}$ to ignore losses on hand pixels (Eq.~\ref{eq:final_loss}), i.e., OA-2DGS,
which produces clean and complete surface meshes.
\Cref{tab:oa_2dgs_ablation} quantitatively compares these methods.

\section{Conclusion}

We advance the research of In-Hand Object Scanning (IHOS) by first introducing a new dataset \toysrus, featuring 4K UHD videos of diverse figurines that have complex geometries.
Since the key challenge for IHOS is accurately estimating camera poses in face of severe dynamic hand occlusions,
we rigorously analyze the well-established Structure-from-Motion (SfM) technique and explain its failure on IHOS videos:
small baselines between consecutive frames, and insufficient feature matches between long-range frames exacerbated by the hand occlusions.
To address these issues,
we propose a simple strategy, called CamStitch, which operates SfM on subsequences of frames sampled at appropriate temporal intervals and stitch the resulting sub-trajectories of estimated camera poses into a global one.
With reliable camera pose trajectory produced by CamStitch,
we further develop occlusion-aware 2D Gaussian Splatting (OA-2DGS) for colored surface reconstruction.
OA-2DGS adopts an opacity loss term, focuses on foreground regions in the reconstruction loss term, and explicitly masks hand regions in loss computation.
Combining CamStitch and OA-2DGS achieves significantly better surface reconstruction results than existing IHOS approaches w.r.t both geometric and photometric metrics.

\section*{Acknowledgments}
The authors thank anonymous reviewers for their constructive feedback, which  substantially improved the manuscript. 
SK gratefully acknowledges support from 
the CK Foundation and the Plato Initiative.
This work is supported by 
the Science and Technology Development Fund of Macau (0058/2025/RIA2, 0067/2024/ITP2),  and
the University of Macau (MYRG-CRG2026-00038-FIC, MYRG-GRG2026-00236-FIC, SRG2023-00044-FST).

\section*{LLM Usage Statement}
This statement discloses the use of Large Language Models (LLMs) and other generative AI tools, collectively referred to as ``AI tools'' hereafter.

Our work includes a study on the performance of modern AI tools, for which we analyze the performance of GPT-6.
Other than this, we used Gemini strictly for language refinement and grammar corrections.
We thoroughly reviewed all AI-assisted revisions to ensure they precisely reflected the original concepts, methodology descriptions, and conclusions of this work.
AI tools were not used to generate research ideas, hypotheses, data, experimental designs, figures, tables, code, or any core technical contributions.

{
    \small
    \bibliographystyle{plainnat}
    \bibliography{arxiv}
}

\clearpage
\setcounter{page}{1}

\renewcommand{\thesection}{\Alph{section}}
\renewcommand{\theHsection}{\Alph{section}}
\setcounter{section}{0}

{
\begin{center}
\Large
\textbf{Appendix}
\end{center}
}

This document supplements our main paper with more details. It is organized as follows: 
\begin{itemize}[leftmargin=15pt, topsep=0pt, itemsep=5pt, parsep=-2pt]
    \item \textbf{Section~\ref{sec:impact_limitation_future}} discusses the societal impact, limitations, and future work.

    \item \textbf{Section~\ref{sec:tech_details_more}} provides more technical details of \ourmethod and compared methods.

    \item \textbf{Section~\ref{sec:toysrus_results_more}} provides more visual results on \toysrus and discusses failure cases.
    
    \item {\bf Section \ref{sec:more_results_more_datasets}} presents more detailed results on more datasets.

    \item {\bf Section \ref{sec:code_dataset_release}} provides code, dataset, intermediate results, and instructions for reproduction.

    \item {\bf Section \ref{sec:impl-3D-Print}} presents extra results of 3D printing based on our reconstructed surface.
\end{itemize}

\section{Impacts, Limitations and Future Work}
\label{sec:impact_limitation_future}

{\bf Societal impacts.} 
Our work advances the research of IHOS.
It helps democratize 3D capture for non-expert common users and potentially facilitate downstream applications such as custom fabrication and 3D printing.
Technically, our work demonstrates why SfM struggles on IHOS video frames,
and presents a novel method to allow SfM shine in IHOS and achieve high-fidelity surface reconstruction on complex figurines.
Serving as a prototype for user-friendly 3D object capture,
our approach empowers everyday users to generate personalized 3D assets, facilitating custom physical fabrications.

{\bf Limitations and Future work.}
We note some limitations of this work.
First,
although \toysrus features complex geometries of figurines, it does not cover many other challenging scenarios, such as textureless, highly reflective, or transparent objects. Future work will include such objects in \toysrus.
Second,
as our method does not specifically handle varying illumination effects (reflections and shadows),
we expect future algorithms to analyze these effects and address potential failure modes caused by them.

\section{More Technical Details}
\label{sec:tech_details_more}

\begin{figure}[h]
\centering
\includegraphics[width=0.6\linewidth]{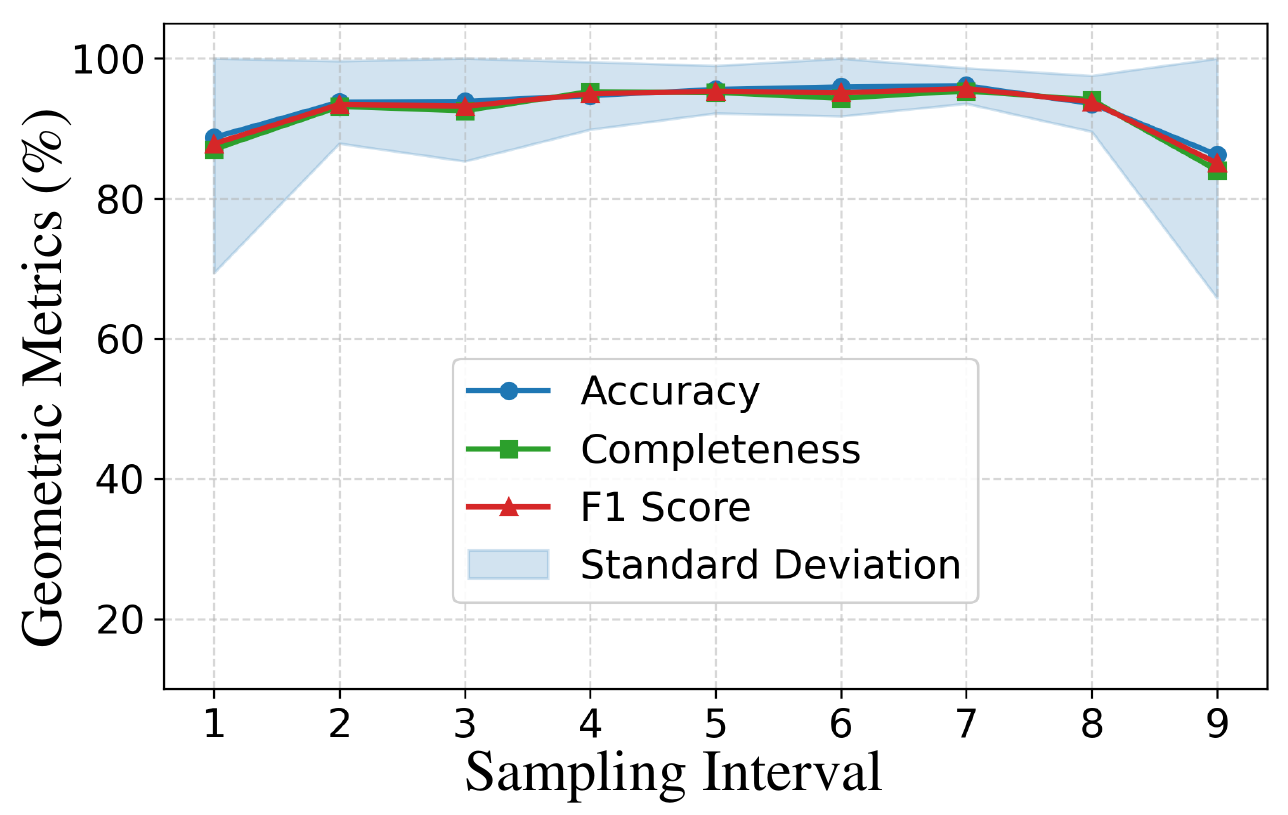}
\vspace{-3mm}
\caption{\small
    \textbf{Sampling interval of CamStitch.}
    We vary the sampling interval to produce subsequences for SfM execution, the first step in CamStitch.
    Too small intervals such as 1 risk CamStitch with the notorious small baseline issue (\cref{fig:baseline_issue});
    too large intervals ($>$8) yield too few feature matches to reliably run SfM, degrading the final performance. Nevertheless, CamStitch is resilient to a large range of sampling intervals.
}
\label{suppl_fig:ablation_study}
\vspace{-1mm}
\end{figure}

\subsection{Analysis of Sampling Interval for CamStitch}

CamStitch implements a divide-and-conquer strategy for SfM execution with the sampling interval as a hyperparameter for sub-sequence construction.
We evaluate CamStitch with different sampling intervals  from 1 to 9.
\Cref{suppl_fig:ablation_study} reports surface reconstruction results using camera poses estimated by CamStitch, in terms of geometric metrics (accuracy, completeness, and F1 score) as well as their min-max ranges.
This demonstrates that CamStitch achieves robust performance across a wide range of sampling intervals.

\subsection{Mathematical Derivation of CamStitch}

Recall that we have a series of camera sub-trajectories along with triangulated 3D points from video sub-sequences. 
To stitch them into a unified coordinate system, we align them sequentially. 
For each pair of sub-sequences, we aim to find a relative transformation matrix $\mathbf{T} \in \text{Sim}(3)$ to jointly map the new camera poses and 3D points into the established coordinate system. 

Note that each sub-trajectory $\mathcal{P}$ comprises camera poses represented by extrinsic matrices $\mathbf{E}$ and triangulated 3D points $\mathbf{P}$.
Using the specific pair of sub-trajectories $\mathcal{P}_1$ and $\mathcal{P}_2$ in \cref{fig:stitch} as an example, our goal is to align and stitch them into a unified trajectory.

First, we establish a pair of corresponding 3D point sets, denoted as $\{\mathbf{P}_1, \mathbf{P}_2\}$, where $\mathbf{P}_1$ and $\mathbf{P}_2$ belong to $\mathcal{P}_1$ and $\mathcal{P}_2$, respectively.
We then estimate $\mathbf{T}$ (consisting of scale $s$, rotation $\mathbf{R}$, and translation $\mathbf{t}$) using the Umeyama algorithm~\cite{umeyama1991least}. 
Specifically, we solve the following optimization problem:
\begin{equation}
\{ s^*, \mathbf{R}^*, \mathbf{t}^* \} = \underset{s, \mathbf{R}, \mathbf{t}}{\arg\min} \sum_{i} \Vert \mathbf{p}_{1}^i - (s\mathbf{R} \mathbf{p}_{2}^i + \mathbf{t}) \Vert_2^2,
\end{equation}
where $\mathbf{p}_{1}^i \in \mathbf{P}_{1}$ and $\mathbf{p}_{2}^i \in \mathbf{P}_{2}$ are the coordinates of the $i$-th corresponding 3D point pair.
By applying $\mathbf{T}$, the 3D points are transformed into the reference frame of $\mathcal{P}_1$ as $\mathbf{P}^{\prime}_{2} = \mathbf{T} \mathbf{P}_{2}$.

Second, we map the camera extrinsics $\mathbf{E}_{2}$ to the reference frame of $\mathcal{P}_1$.
Since the relative spatial relationship between the cameras and 3D points is preserved during this transformation, the 3D point coordinates observed in the local camera coordinate system, $\mathbf{P}_{cam}$, remain invariant.
Thus, we have:
\begin{equation}
\mathbf{P}_{cam} = \mathbf{E}^{\prime}_{2} \mathbf{P}^{\prime}_{2} = \mathbf{E}_{2} \mathbf{P}_{2},
\end{equation}
where $\mathbf{E}^{\prime}_{2}$ represents the camera extrinsics after transformation.
By substituting $\mathbf{P}_{2} = \mathbf{T}^{-1} \mathbf{P}^{\prime}_{2}$ into the equation above, we obtain:
\begin{equation}
\mathbf{P}_{cam} = \mathbf{E}^{\prime}_{2} \mathbf{P}^{\prime}_{2} = \mathbf{E}_{2} \mathbf{T}^{-1} \mathbf{P}^{\prime}_{2}.
\end{equation}
By equating the transformation matrices applied to $\mathbf{P}^{\prime}_{2}$, it follows that $\mathbf{E}^{\prime}_{2} = \mathbf{E}_{2} \mathbf{T}^{-1}$.
Thus, the updated 3D points and camera extrinsics can be jointly summarized as:
\begin{equation}
\mathbf{P}^{\prime}_{2} = \mathbf{T} \mathbf{P}_{2}, \quad \mathbf{E}^{\prime}_{2} = \mathbf{E}_{2} \mathbf{T}^{-1}.
\end{equation}
Intuitively, this implies that when the 3D points are transformed by $\mathbf{T}$, the camera extrinsics are updated via the inverse transformation $\mathbf{T}^{-1}$ to ensure the camera observations remain geometrically consistent.

Finally, the two sets of triangulated 3D points and camera extrinsics are aligned and unified, yielding the stitched trajectory $\mathcal{P}_{1} \cup \mathcal{P}^{\prime}_{2} = \{\mathbf{P}_{1} \cup \mathbf{P}^{\prime}_{2}, \mathbf{E}_{1} \cup \mathbf{E}^{\prime}_{2}\}$.
By applying this method iteratively, we stitch all sub-trajectories into a unified global camera trajectory.

\begin{table*}[t]
\small
    \centering
    \caption{Compared with existing datasets,
    \toysrus better meets the rising demand for custom and high-resolution 3D capture.
    It features figurines that have complex geometries (\cref{fig:dataset}),
    and 4K UHD videos captured by a hand-held smartphone in a user-friendly manner.
    }
    \label{suppl_tab:dataset}
    \vspace{-3mm}
    \setlength{\tabcolsep}{1.mm}
    \resizebox{1\textwidth}{!}{
    \small
    \begin{tabular}{l c c c c c c c c c c c c c c c  c c c c c c}
    \toprule
        dataset 
        & task
        & publicly 
        & hand-held
        & frame 
        & object 
        & \#frames 
        & firm-grasp 
        & customized
        & full video
        & \#objects  
        \\
        & & available?   
        & smartphone?
        & res. 
        & res. 
        & /object 
        & vs. toying 
        & objects?
        & release?
        &
        \\
    \midrule

        HOD~\cite{huang2022reconstructing} 
        & {\color{darkgreen} IHOS} 
        & \xmark 
        & \color{lightred}No
        & {\color{darkgreen}4K} 
        & \color{darkgreen}100$\sim$1800
        & \color{darkgreen}$\sim$1800
        & \color{lightred}firm
        & \color{darkgreen}Yes
        & \color{lightred}No
        & 35 
        \\ 
        
        HOPE~\cite{tyree20226} 
        & {\color{lightred}pose est.} 
        & \cmark 
        & \color{lightred}No
        & {\color{lightred} 1080p} 
        & \color{lightred}100$\sim$350
        & \color{lightred}$\sim$300
        & \color{darkgreen}toying
        & \color{lightred}No
        & \color{lightred}No
        & 28 
        \\

         HO3D~\cite{hampali2020honnotate} 
         & {\color{lightred} pose est.} 
         & \cmark 
        & \color{lightred}No
        & {\color{lightred}480p} & \color{lightred}100$\sim$300
        & \color{darkgreen}$\sim$1500
        & \color{lightred}\  firm
        & \color{lightred}No
        & \color{darkgreen}Yes
         & 10 
        \\
        
        IHO~\cite{zhong2024color} 
        & {\color{darkgreen}IHOS} 
        & \cmark 
        & \color{lightred}No
        & {\color{lightred} 1080p}
        & \color{darkgreen}300$\sim$1,000 
        & \color{darkgreen}$\sim$2000
        & \color{darkgreen}toying
        & \color{lightred}No
        & \color{darkgreen}Yes
        & 4 
        \\
        
        \toysrus 
        & {\color{darkgreen}IHOS} 
        & \cmark 
        & \color{darkgreen}Yes
        & {\color{darkgreen}4K} 
        & \color{darkgreen}300$\sim$3,000 
        & \color{darkgreen}$\sim$2000
        & \color{darkgreen}toying
        & \color{darkgreen}Yes
        & \color{darkgreen}Yes
        & \textbf{40}
        \\
    \bottomrule
    \end{tabular}
    }
    \vspace{-1mm}
\end{table*}

\begin{figure}[t]
\centering
\includegraphics[width=0.6\linewidth]{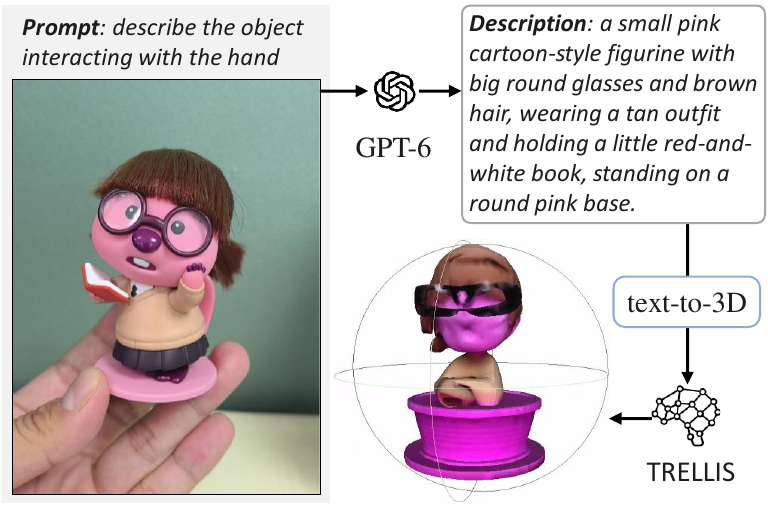}
\vspace{-2mm}
\caption{\small
    \textbf{Failure modes of Dynhor~\cite{jiang2025hand}.} 
    As the foundation models cannot reliably handle uncommon figurines, Dynhor performs poorly on \toysrus.}
\label{suppl_fig:Dynhor_repro}
\vspace{-0mm}
\end{figure}

\subsection{Re-Implementation of Dynhor}

The original Dynhor pipeline~\cite{jiang2024hand} prompts ChatGPT to generate a text description for the hand-held object in the video, which is subsequently fed into the text-to-3D model Luma to generate a 3D proxy mesh. 
However, the official source code for Dynhor is partially unavailable, and the Luma foundation model has been deprecated. 
Therefore, we re-implement this pipeline using GPT-5.2~\cite{singh2025openai} for text generation and the more recent text-to-3D model TRELLIS~\cite{Xiang_2025_CVPR} for mesh generation. 
Furthermore, since their proposed joint optimization stage remains unreleased, our comparison is directly based on the 3D meshes generated by the text-to-3D stage. 
An example of the detailed text prompt template and the corresponding generated description is provided in \cref{suppl_fig:Dynhor_repro}. 
As observed, Dynhor fails to accurately capture the appearance of objects with custom and complex geometries in \toysrus.

\section{More Evaluations on \toysrus}
\label{sec:toysrus_results_more}

\begin{figure*}[!htbp]
    \centering
    \includegraphics[width=0.95\textwidth]{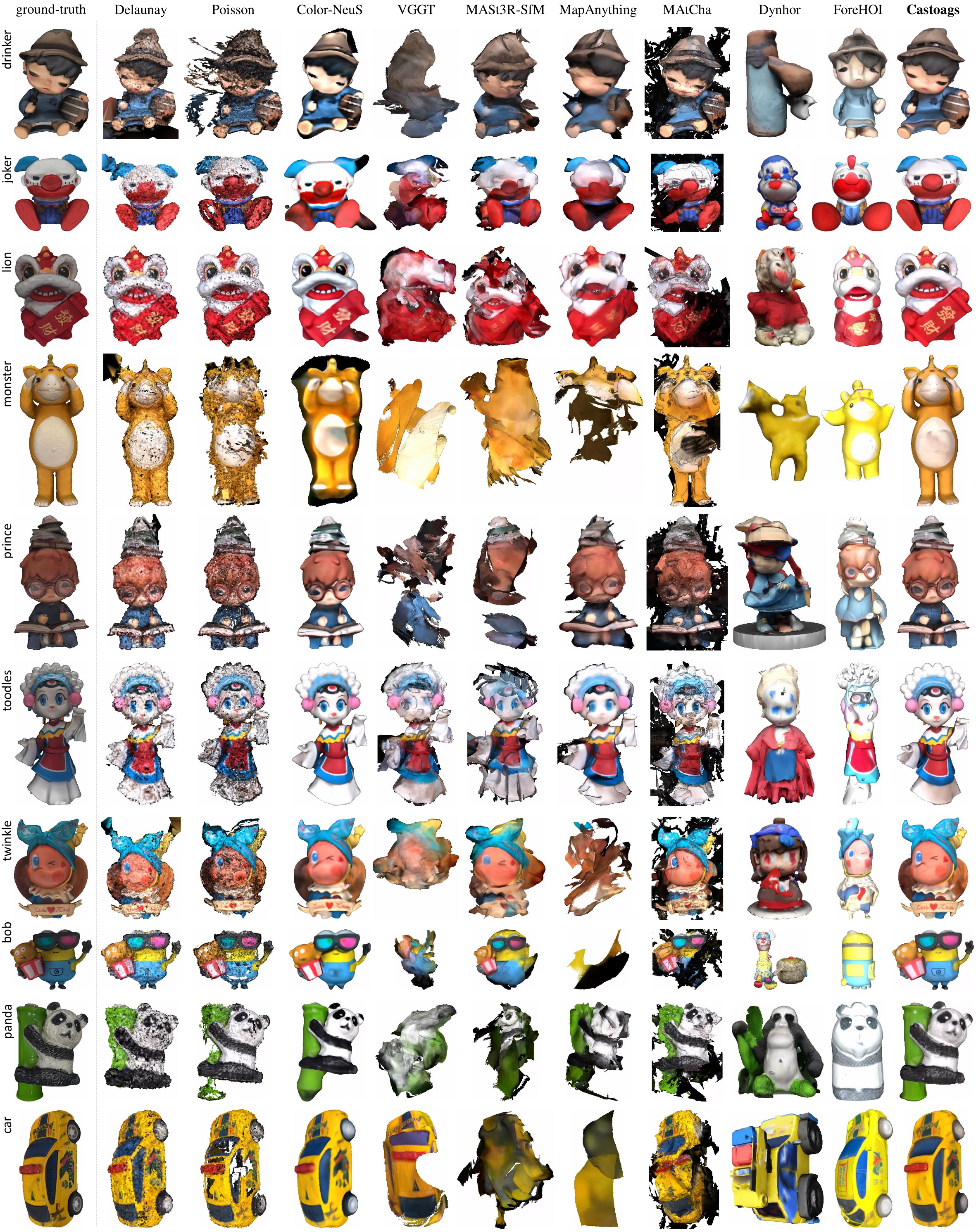}
    \vspace{-3mm}
    \caption{\textbf{Extensive visualizations on the \toysrus dataset.} We visually present 10 additional reconstructed results corresponding to the quantitative metrics in \cref{tab:toysrus_result}. 
    While Color-NeuS~\cite{zhong2024color} achieves comparable visual quality, it requires $\sim$12 hours per object, which severely hinders its practical application in IHOS. 
    In contrast, our approach delivers highly competitive visual fidelity in merely $\sim$20 minutes, demonstrating a massive leap in efficiency.}
    \label{suppl_fig:toysrus_result_more}
\end{figure*}

\textbf{Detailed Statistics of \toysrus.}
To comprehensively compare \toysrus with existing datasets, we provide a detailed summary in \cref{suppl_tab:dataset}. 
For example, \textit{object res.} denotes the actual pixel area occupied by the targeted object within the full \textit{frame res.},
reflecting the fact that high object resolution is crucial for achieving superior reconstruction quality.
Compared to other datasets,
our \toysrus provides 4K UHD videos and ensures that the object occupies a significant portion of the image (up to 3,000 pixels). 
Another column \textit{\#frames /object} indicates the total number of frames available for camera pose estimation and surface reconstruction. 
Datasets like HOPE~\cite{tyree20226} only release a limited subset of sparsely sampled frames (e.g., $\sim$300) rather than the full continuous video. 
This inherent data sparsity causes a significant gap compared to data collection in real-world scenarios. 
Consequently, when methods repurpose such datasets for the IHOS task (e.g., Dynhor~\cite{jiang2025hand}), they are forced to utilize 3D proxy generation to compensate for the lack of sufficient video frames and object information.
However, this proxy-based reliance ultimately fails to reconstruct the authentic geometric shapes of uncommon objects.
Notably, our \toysrus is explicitly tailored for IHOS.
By recording full videos with everyday smartphones to cover all exposed object surfaces, it closely aligns with real-world settings, thereby better facilitating methodological advancements in this field.

\textbf{More Visual Results on \toysrus.}
We present more visual results on the \toysrus, illustrated in \cref{suppl_fig:toysrus_result_more}. 
While Color-NeuS~\cite{zhong2024color} achieves visually comparable reconstructions overall, a closer inspection reveals that it often struggles with fine structures and exhibits geometric artifacts. 
As highlighted in \cref{suppl_fig:toysrus_result_more}, our method successfully recovers these intricate geometric details with much higher fidelity than Color-NeuS. 
More importantly, relying on implicit representations, Color-NeuS requires $\sim$12 hours of optimization per object.
In contrast, our approach leverages highly efficient explicit representations, delivering these high-fidelity surfaces in merely $\sim$20 minutes. 
This demonstrates a clear advantage in practical efficiency without compromising reconstruction quality.

\begin{figure*}[!htbp]
    \centering
    \includegraphics[width=0.99\linewidth]{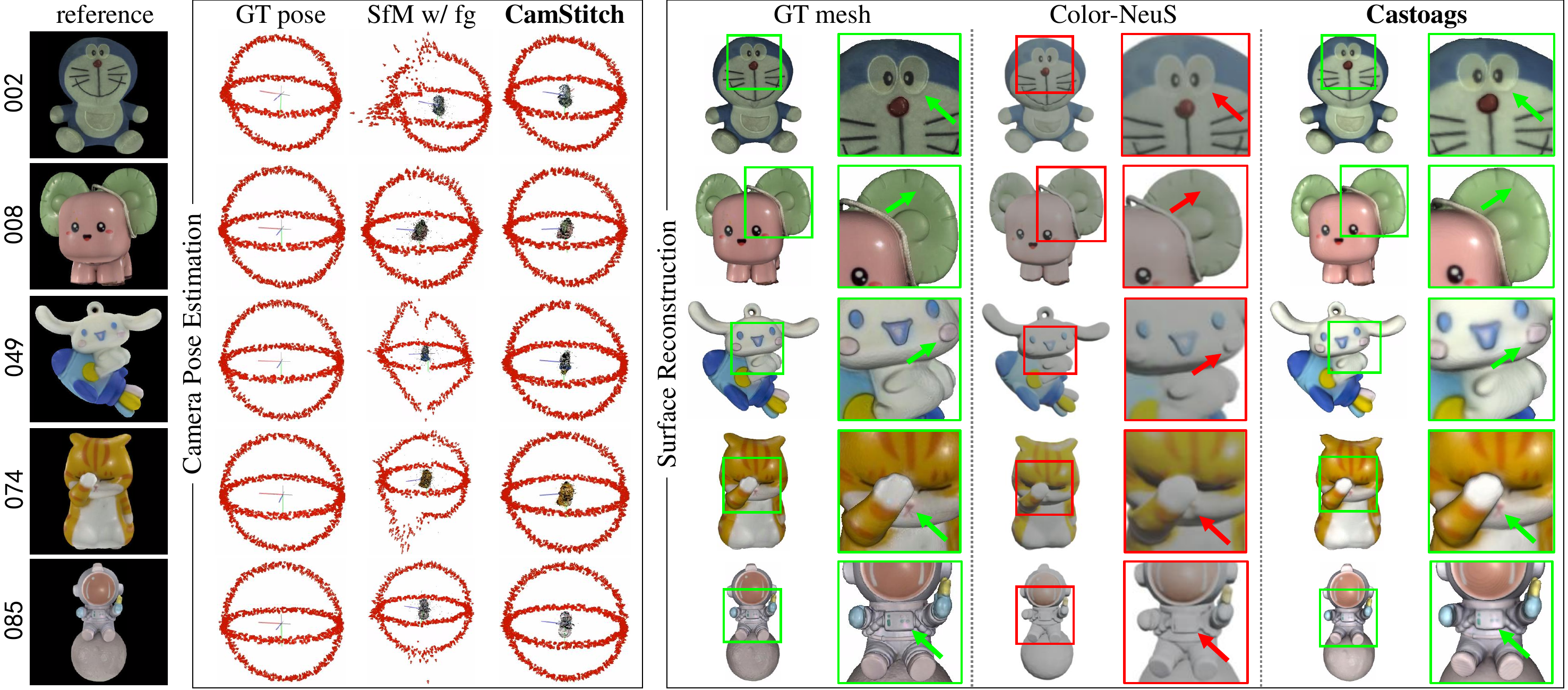}
    \vspace{-2.5mm}
    \caption{\small \textbf{Qualitative results on OmniObject3D.} We evaluate camera pose estimation (left) and surface reconstruction (right). CamStitch yields camera trajectories aligned with the ground truth (GT pose), resolving track drift in naive SfM (SfM w/ fg). Castoags produces high-fidelity surfaces compared to Color-NeuS~\cite{zhong2024color}, preserving finer geometric details consistent with the ground truth (GT mesh). We also provide quantitative results on camera pose estimation (see \cref{suppl_tab:campose_estimation}) and surface reconstruction (see \cref{suppl_tab:oo3d_result}).}
    \label{suppl_fig:oo3d_result}
\end{figure*}

\begin{table*}[t]
    \centering
    \begin{minipage}[t]{0.48\linewidth}
        \centering
        \setlength{\tabcolsep}{1.5mm}
        \caption{\small \textbf{Quantitative results of camera pose estimation.} We compare CamStitch against naive SfM (SfM w/ fg) on OmniObject3D. CamStitch yields higher pose accuracy across all metrics, requiring less processing time.}
        \vspace{-3mm}
        \resizebox{0.8\linewidth}{!}{
        \begin{tabular}{lrrrrr}
            \toprule
            method 
            & mRE$\downarrow$ & ATE$\downarrow$ & RPE$_r$$\downarrow$ & RPE$_t$$\downarrow$ & time$\downarrow$ \\
            \midrule
            SfM w/ fg & 1.26 & 29.46 & 21.65 & 0.61 & 0.43  \\
            \rowcolor{col33}{\textbf{CamStitch}}& \textbf{0.72} & \textbf{3.32} & \textbf{2.02} & \textbf{0.41} & \textbf{0.16} \\
            \bottomrule
        \end{tabular}}
        \label{suppl_tab:campose_estimation}
    \end{minipage}\hfill
    \begin{minipage}[t]{0.48\linewidth}
        \centering
        \setlength{\tabcolsep}{3mm}
        \caption{\small \textbf{Quantitative results of surface reconstruction.} We compare the geometric accuracy of Castoags against Color-NeuS using Chamfer Distance on OmniObject3D. Castoags achieves lower errors across all test objects.
        }
        \vspace{-3mm}
        \resizebox{0.85\linewidth}{!}{
        \begin{tabular}{lccccc}
            \toprule
            method & {\tt 002} & {\tt 008} & {\tt 049} & {\tt 074} & {\tt 085}\\
            \midrule
            Color-NeuS  & \underline{0.27} & \underline{1.91} & \underline{0.53} & \underline{0.40} & \underline{0.74} \\
            
            \rowcolor{col33}{\bf \ourmethod} & \textbf{0.11} & \textbf{0.45} & \textbf{0.11} & \textbf{0.35} & \textbf{0.08} \\
            \bottomrule
        \end{tabular}}
        \label{suppl_tab:oo3d_result}
    \end{minipage}
    \vspace{-0mm}
\end{table*}

\begin{figure}[t]
\centering
\includegraphics[width=0.75\linewidth]{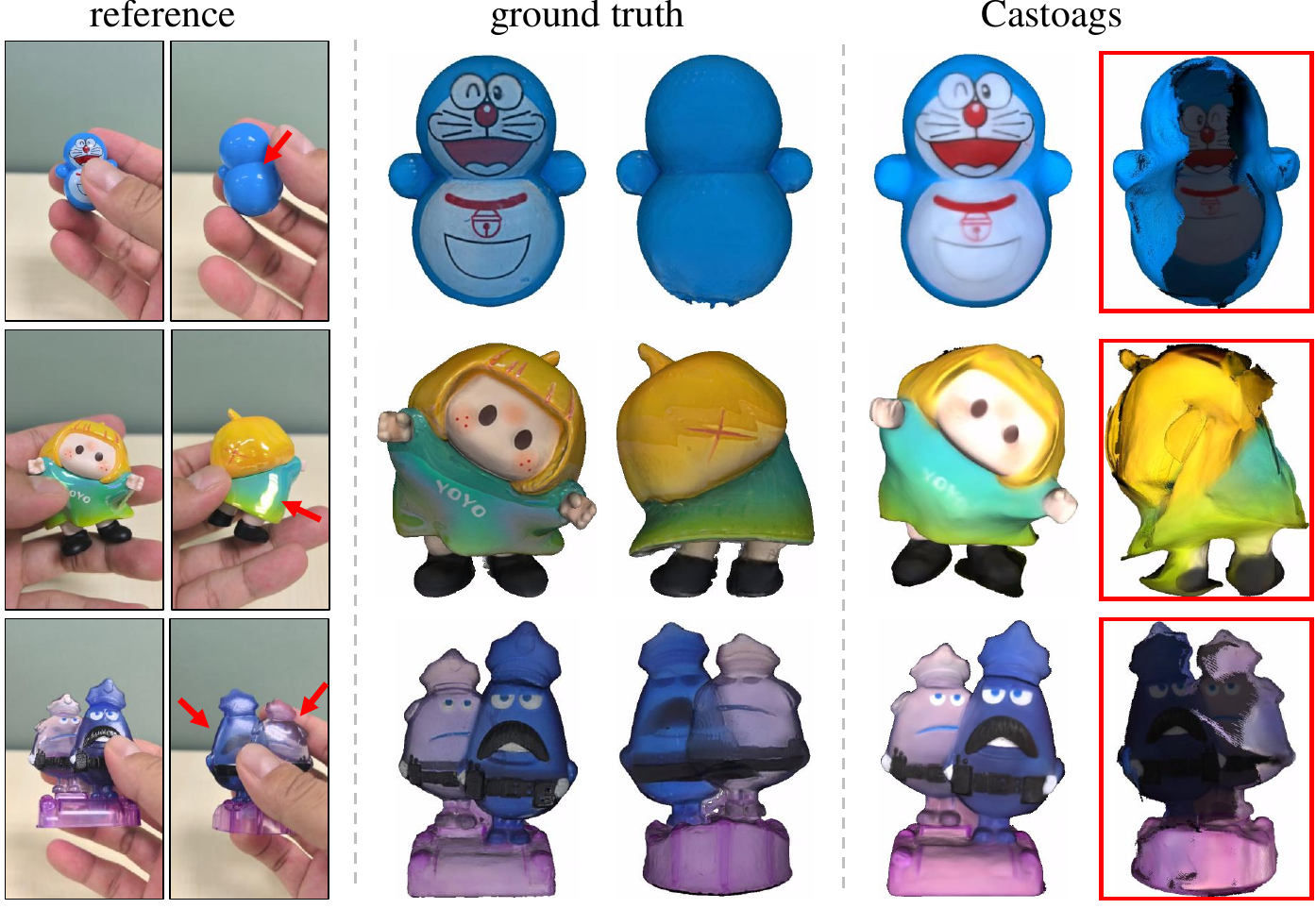}
\vspace{-2mm}
    \caption{\small \textbf{Failure cases.} 
    \ourmethod reconstructs incomplete mesh surfaces under specific conditions: single-color textures, transparent materials, and smooth surfaces.}
\label{suppl_fig:failure_cases}
\vspace{-0mm}
\end{figure}

\textbf{Failure Case Analysis.} 
Our method also encounters challenges, as the \toysrus dataset contains diverse customized objects across various materials, colors, and geometries. 
We visualize several representative failure cases in \cref{suppl_fig:failure_cases} to illustrate these limitations and present future directions. 
In practice, smooth surfaces and transparent materials frequently co-occur, causing specular reflections under changing illumination, where highlight spots (indicated by red arrows in \cref{suppl_fig:failure_cases}) obscure the true colors of the objects. 
Single-color textures lack trackable features, hindering feature detection and matching in CamStitch. 
Subsequently, this inaccurate pose initialization further impairs surface reconstruction in OA-2DGS due to the resulting flawed camera trajectories.

\section{More Detailed Results on More Datasets}
\label{sec:more_results_more_datasets}

\textbf{Benchmarking on OmniObject3D.} 
We further evaluate our method on the OmniObject3D~\cite{wu2023omniobject3d} dataset to demonstrate its generalizability in object reconstruction. 
OmniObject3D provides ground-truth meshes but features only semi-spherical camera coverage; thus, we generate novel camera trajectories around the objects and render images to simulate the IHOS setting. 
Specifically, we select 5 representative objects as shown in~\cref{suppl_fig:oo3d_result}.
To separately assess camera pose accuracy and efficiency, we first employ standard tracking metrics, including mRE (px), ATE (m), $\text{RPE}_t$ (m), and $\text{RPE}_r$ ($^\circ$), along with running time (hours), to evaluate our CamStitch against naive SfM.
As detailed in~\cref{suppl_tab:campose_estimation}, the proposed CamStitch provides a highly reliable and efficient camera initialization. 
Subsequently, we use Chamfer Distance (CD) to evaluate the surface reconstruction quality of \ourmethod. 
As shown in~\cref{suppl_tab:oo3d_result}, \ourmethod achieves higher geometric accuracy and color fidelity compared to Color-NeuS~\cite{zhong2024color}.
These quantitative gains are supported by the qualitative results in \cref{suppl_fig:oo3d_result}, where our pipeline successfully recovers realistic appearances and fine surface details.

\section{Code and Dataset Release}
\label{sec:code_dataset_release}

We include our code as part of the supplementary material.
We do not include \toysrus because a single video is more than 300MB, totaling over 11GB for all 40 objects.

{\bf Requirements.} Running the code requires Python, PyTorch, COLMAP, and other common packages.
We provide setup instructions in \textit{readme.txt} and list all dependencies in the \textit{requirements.txt} file for convenient installation.
Below are the versions of Python, PyTorch, and COLMAP toolbox used in this work:
\begin{itemize}
    \item COLMAP: 3.13.0.dev0
    \item Python: 3.8.18
    \item PyTorch: 2.0.0
\end{itemize}

{\bf License.} We will release our code under the MIT license.

{\bf Demo.} We provide detailed instructions for reproducing our experiments in the following Jupyter Notebook files:
\begin{itemize}
    \item \textit{step1\_pose\_estimation.ipynb} contains the implementation of \textbf{\ourmethod} for camera pose estimation.
    \item \textit{step2\_surf\_reconstruction.ipynb} contains the implementation of \textbf{OA-2DGS} for surface reconstruction.
    \item \textit{step3\_surf\_evaluation.ipynb} contains the implementation of geometric and photometric evaluation.
\end{itemize}

{\bf Results.} We provide the reconstruction results of a sample object in \toysrus.
The estimated camera trajectories are saved in the standard COLMAP format under the \textit{sparse/0/} directory, which can be directly opened and visualized using the COLMAP GUI. 
Additionally, we provide the reconstructed surface mesh named \textit{loopy.ply}, which can be visualized using MeshLab and online PLY visualizer.

\section{Quality Examination via 3D Printing}
\label{sec:impl-3D-Print}

\begin{figure}[t]
    \centering
    \includegraphics[width=0.85\linewidth]{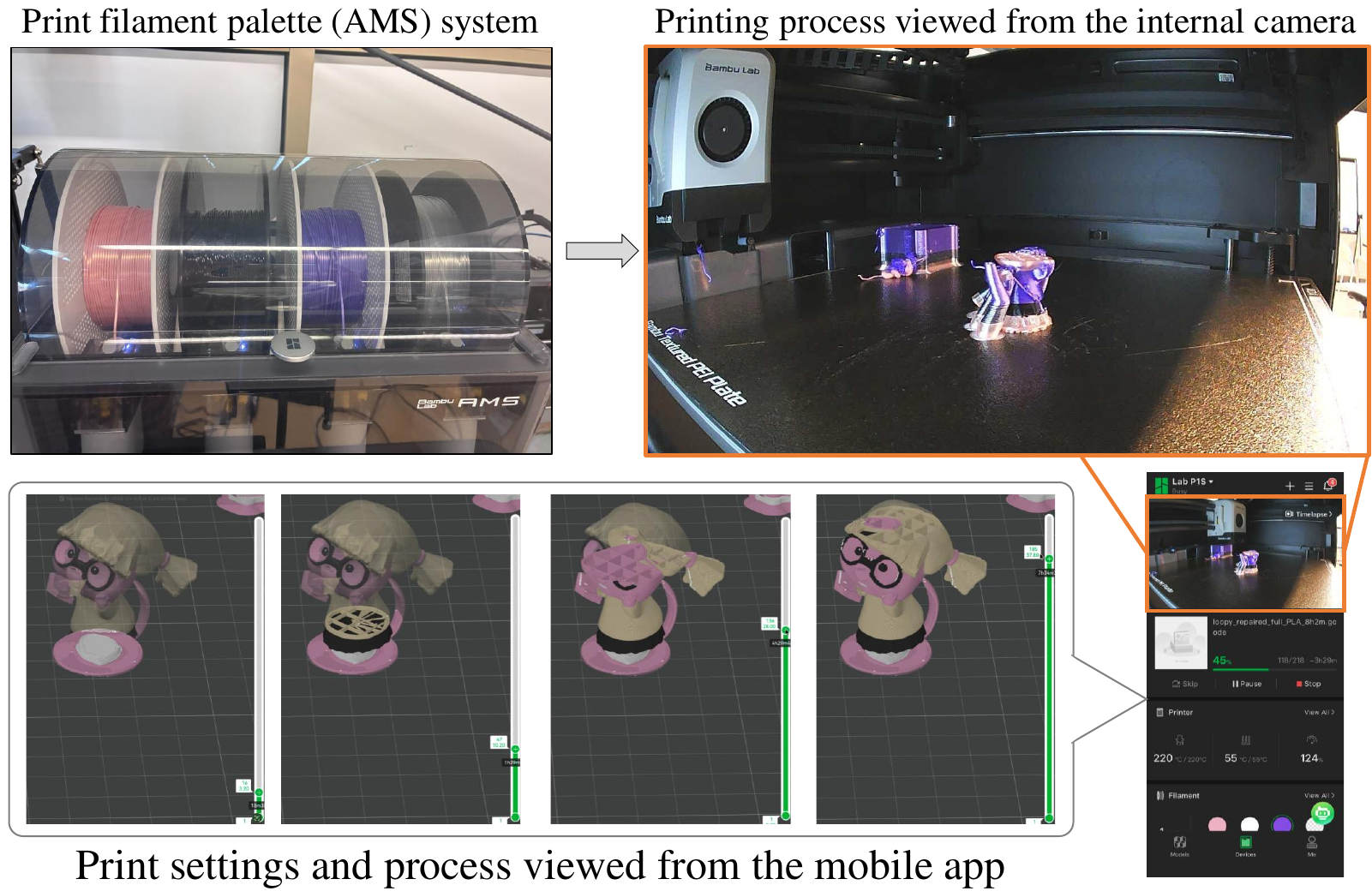} 
    \vspace{-1mm}
    \caption{
    \textbf{Illustration of 3D printing.} 
    To further examine the quality of reconstructed surfaces, we turn to 3D printing to physically print the object based on its reconstructed surface.    
    In brief, we use a Bambu Lab multi-color printer equipped with the Automated Material System (AMS).
    An internal camera in the printer helps monitor printing progress, which can be view in mobile app.
    }
    \label{suppl_fig:printing}
\end{figure}

While our proposed evaluation protocol does not require physical fabrication to assess surface reconstruction quality,
we perform 3D printing to validate the practical utility of our IHOS method for custom fabrication.
\cref{suppl_fig:printing} illustrates the 3D printing workflow.
Below, we detail specific steps involved in our commodity desktop 3D printing pipeline.
We also include a video named \textit{printing.mp4} in the supplementary materials that records the printing process.

\begin{figure}[t]
  \centering
    \includegraphics[width=0.75\linewidth]{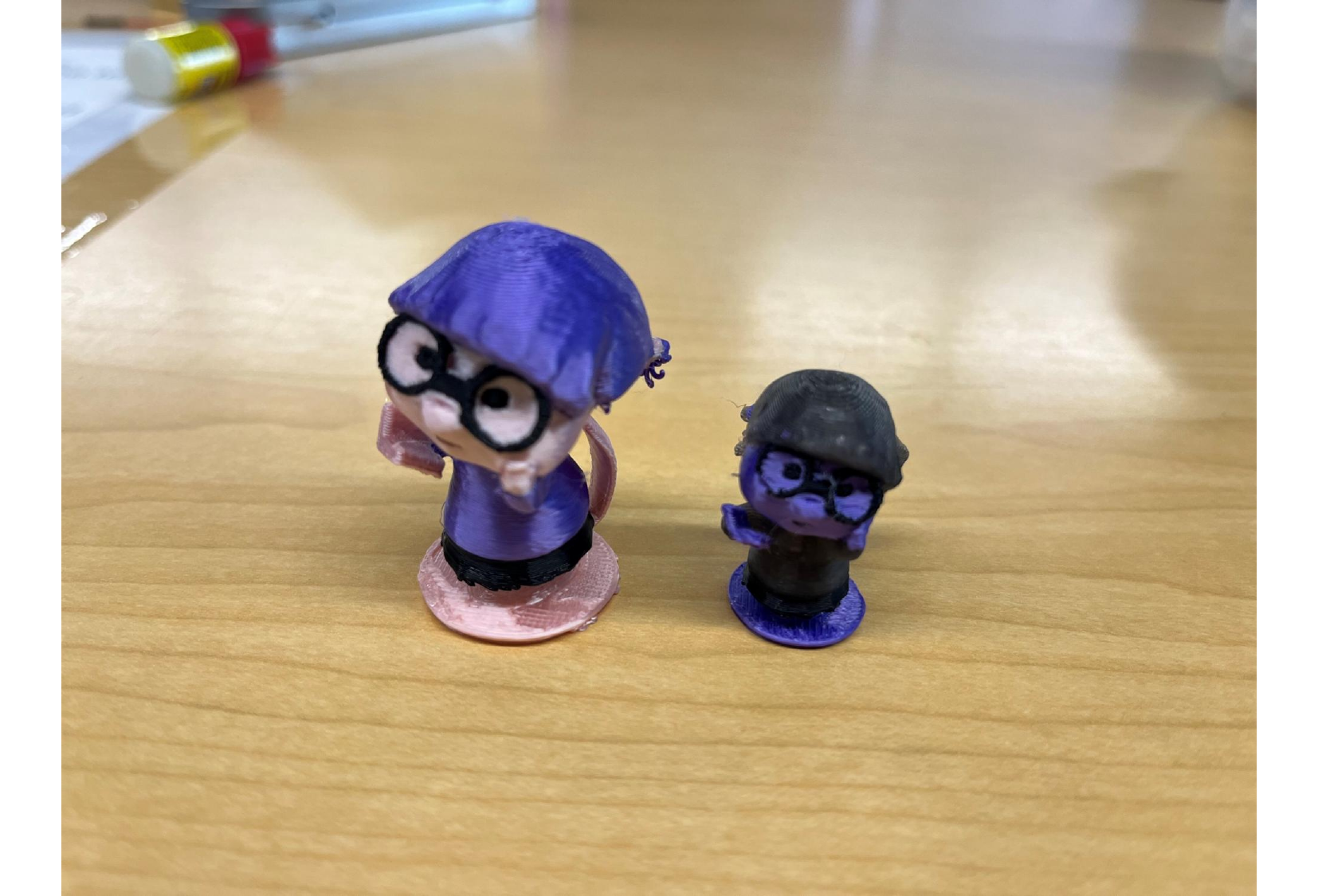}
    \small
    \vspace{-2mm}
    \caption{\small
    We change filament and 3D print the same object at a large and a small scale, respectively.
    At a small scale, 
    we see a failure on the pigtail because of minimal overhang that require a little to no supporting structure underneath, which is automatically generated by the slicer.    
    Nevertheless, this demonstrates that the reconstructed colored surfaces by our method serve as a base model for custom fabrication and 3D printing.
    }
    \label{suppl_fig:3Dprinting-filament}
    \vspace{-0mm}
\end{figure}

{\bf Full color printing.}
We fabricate the objects using a Bambu Lab multi-color printer equipped with the Automatic Material System (AMS). 
To support Bambu Studio slicer,
we save the reconstructed colored mesh in .stl format,
a universal file format for 3D printing.
We perform mesh repairs to ensure watertight geometry, which is required by 3D printing.
The slicer automatically clusters color schemes, mapping the original full-color surface into a limited palette of four filaments. 
The models are sliced and printed using standard settings:
0.2 mm standard layer height, 0.4 mm nozzle, 15\% standard infill, conventional PLA filaments, and printing speeds between 200-300 mm/s.

{\bf Remarks.}
Executing this workflow yields several insights.
First, the reconstructed surfaces provide a solid foundation for fabrication, yet additional manual intervention is required to make them print-ready.
For example, we manually repair specific mesh regions to ensure watertight geometry necessary for successful 3D printing. 
Second, 
achieving finer color segmentation is difficult.
Manually selecting detailed color regions necessitates frequent filament exchanges. This results in significant material waste due to purging, nozzle clearance, and refilling cycles.
Third, the choice of filament and printing scale influence the quality of 3D printed object. For example,
printing at a small scale can fail to 3D print the pigtails of {\tt loopy} as shown in \cref{suppl_fig:3Dprinting-filament}.

\end{document}